\documentclass{article}

\usepackage{microtype}
\usepackage{graphicx}
\usepackage{subfigure}
\usepackage{booktabs} 
\usepackage{multirow}

\usepackage{hyperref}

\newcommand{\sgens}[1]{\mathbb{R}^{#1}}

\newcommand{\op}[1]{\operatorname{#1}}

\usepackage[accepted]{icml2024}

\usepackage{amsmath}
\usepackage{amssymb}
\usepackage{mathtools}
\usepackage{amsthm}

\usepackage[capitalize,noabbrev]{cleveref}

\theoremstyle{plain}
\newtheorem{theorem}{Theorem}[section]
\newtheorem{proposition}[theorem]{Proposition}
\newtheorem{lemma}[theorem]{Lemma}
\newtheorem{corollary}[theorem]{Corollary}
\theoremstyle{definition}
\newtheorem{definition}[theorem]{Definition}

\theoremstyle{remark}
\newtheorem{remark}[theorem]{Remark}

\usepackage[textsize=tiny]{todonotes}

\icmltitlerunning{Positive Concave Deep Equilibrium Models}

\begin{document}

\twocolumn[
\icmltitle{Positive Concave Deep Equilibrium Models}



\icmlsetsymbol{equal}{*}

\begin{icmlauthorlist}
\icmlauthor{Mateusz Gabor}{yyy}
\icmlauthor{Tomasz Piotrowski}{comp}
\icmlauthor{Renato L. G. Cavalcante}{sch}
\end{icmlauthorlist}

\icmlaffiliation{yyy}{Faculty of Electronics, Photonics, and Microsystems,Wrocław University of Science and Technology, Wrocław, Poland}
\icmlaffiliation{comp}{Faculty of Physics, Astronomy and Informatics, Nicolaus Copernicus University, Toruń, Poland}
\icmlaffiliation{sch}{Fraunhofer Heinrich Hertz Institute, Berlin, Germany}

\icmlcorrespondingauthor{Mateusz Gabor}{mateusz.gabor@pwr.edu.pl}

\icmlkeywords{Machine Learning, ICML}

\vskip 0.3in
]



\printAffiliationsAndNotice{}  

\begin{abstract}
Deep equilibrium (DEQ) models are widely recognized as a memory efficient alternative to standard neural networks, achieving state-of-the-art performance in language modeling and computer vision tasks. These models solve a fixed point equation instead of explicitly computing the output, which sets them apart from standard neural networks. However, existing DEQ models often lack formal guarantees of the existence and uniqueness of the fixed point, and the convergence of the numerical scheme used for computing the fixed point is not formally established. As a result, DEQ models are potentially unstable in practice. To address these drawbacks, we introduce a novel class of DEQ models called positive concave deep equilibrium (pcDEQ) models. Our approach, which is based on nonlinear Perron-Frobenius theory, enforces nonnegative weights and activation functions that are concave on the positive orthant. By imposing these constraints, we can easily ensure the existence and uniqueness of the fixed point without relying on additional complex assumptions commonly found in the DEQ literature, such as those based on monotone operator theory in convex analysis. Furthermore, the fixed point can be computed with the standard fixed point algorithm, and we provide theoretical guarantees of its geometric convergence, which, in particular, simplifies the training process. Experiments demonstrate the competitiveness of our pcDEQ models against other implicit models.
\end{abstract}

\section{Introduction}
Implicit models \cite{bai2019deep,bai2020multiscale,chen2018neural,el2021implicit,baker2023implicit,tsuchida2023deep,revay2020lipschitz,wei2021certified,geng2021training} are attracting considerable interest owing to their improved memory efficiency compared to standard neural networks. These models solve implicit equations instead of explicitly computing the output of the layers, and they can be divided into two main categories: neural ordinary differential equations (Neural ODEs) \cite{chen2018neural} and deep equilibrium models (DEQ) models \cite{bai2019deep}. 

Neural ODEs solve differential equations parameterized by the neural network input, with the output representing the solution to these equations. On the other hand, the implicit layers of DEQ models solve nonlinear fixed point equations that are not necessarily derived from differential equations.  An interesting aspect about DEQ models is that a single implicit DEQ layer emulates a standard neural network with an infinite number of layers and tied weights. While both DEQ models and neural ODEs require constant training memory, DEQ models often outperform neural ODEs, achieving state-of-the-art results in language modeling tasks \cite{bai2019deep} and computer vision tasks \cite{bai2020multiscale,xie2022optimization}.  However, a potential limitation of standard DEQ models is that they are based on iterative methods that require careful initialization, tuning of hyperparameters, and special regularization (e.g., recurrent dropout) to ensure convergence to the fixed point, which is hard to guarantee with existing approaches. In general, standard DEQ models operate heuristically, and in many cases they even lack formal guarantees regarding the existence and uniqueness of the fixed point problem being solved.

To overcome the limitations of existing DEQ models, we introduce a new variant called positive concave deep equilibrium (pcDEQ) models. In particular, these pcDEQ models address issues related to the existence and uniqueness of the fixed point. Furthermore, the fixed points of the proposed pcDEQ models can be easily computed with the standard fixed point iteration, and we provide formal guarantees of geometric convergence. The theoretical foundation of pcDEQ models is rooted in nonlinear Perron-Frobenius (NPF) theory \cite{lemmens2012nonlinear,lins2023unified}, which is commonly used in the analysis of nonnegative monotonic (order-preserving) and scalable functions. To ensure these properties in pcDEQ models, we enforce nonnegative weights and activation functions that are concave in the nonnegative orthant. An additional advantage of our proposed model is that the standard Jacobian-based backpropagation algorithm \cite{bai2019deep} can be used for training without requiring any changes.

We summarize our contributions as follows:
\vspace{-0.5\baselineskip}
\begin{enumerate}
    \item We propose a new class of DEQ models, called pcDEQ models, which are based on nonlinear Perron-Frobenius theory and are equipped with guarantees of the existence and uniqueness of the fixed point. Furthermore, we prove that the standard fixed point iteration for the proposed pcDEQ models converges to the fixed point geometrically fast.
    \item We empirically show that, for the proposed pcDEQ architectures, in practice, only a few iterations are needed to achieve numerical convergence, and the number of iterations does not increase over the course of training.
    \item We demonstrate competitive improvement of the proposed approach in terms of accuracy and number of parameters over existing alternatives for image classification tasks.
\end{enumerate}

\vspace{-\baselineskip}
\section{Related Work}
In the seminal paper \cite{bai2019deep}, DEQ models have been applied to language modeling tasks, and they have been shown to outperform standard neural networks constructed with a similar number of parameters. In subsequent studies, Bai \textit{et al.} proposed \textit{multiscale} extensions of DEQ models \cite{bai2020multiscale}, where a single DEQ model is used for image classification and image segmentation. Since these pioneering studies, DEQ models have been successful in many applications, including, to name a few, object detection \cite{wang2023deep}, optical flow estimation \cite{bai2022deep}, video semantic segmentation \cite{ertenli2022streaming}, medical image segmentation \cite{zhang2022efficient}, snapshot compressive imaging \cite{zhao2023deep}, image denoising \cite{chen2023equilibrium,gkillas2023connections}, machine translation \cite{zheng2023deep}, inverse problems \cite{gilton2021deep,zou2023deep}, music source separation \cite{koyama2022music}, federated learning \cite{gkillas2023deep}, and diffusion models \cite{geng2023one,pokle2022deep}. These models are strongly based on the fixed point theory that we summarize below.




Let $\sgens{n}$ be the standard Euclidean metric space and let $f: \sgens{n}\to\sgens{n}$. If $f$ is a Lipschitz contraction, meaning it is Lipschitz continuous with a Lipschitz constant $L<1$, then the Banach fixed point theorem guarantees the uniqueness and existence of the fixed point $x^\star=f(x^\star)$ in $\sgens{n}$. Moreover, the fixed point iteration $x_{k+1}=f(x_k)$ converges linearly to $x^\star$ for any initial point $x_1\in\sgens{n}$. However, if $f$ is not a Lipschitz contraction, questions related to the existence and uniqueness of the unique point, and also the convergence of the fixed point iteration, become more delicate, and these issues have been the subject of extensive research in the mathematical literature.

In modern convex analysis, there has been a significant focus on nonexpansive mappings in Hilbert spaces \cite{yamada2011minimizing}, which are mappings with Lipschitz constant equal to one, and their relation to monotone operator theory \cite{baus17}. Nonexpansive mappings do not necessarily have a fixed point, and, if the fixed point set is nonempty, it may not be a singleton in general. Furthermore, even if the fixed point exists, then the fixed point iteration may fail to converge, in which case one can resort to various iterative methods based on the the Krasnosel'skii-Mann iteration to ensure it. Particular instances of this iteration include well-known algorithms used in machine learning, such as the projected gradient method, the proximal forward-backward splitting method, the Douglas-Rachford splitting method, the projection onto convex sets method, and many others \cite{yamada2011minimizing}.

 Monotone operator theory in DEQ models has been explored in \cite{winston2020monotone}, a study that introduces the monotone operator deep equilibrium (monDEQ) models. These models ensure both the existence and uniqueness of the fixed point, while also guaranteeing the convergence of the forward-backward splitting algorithm and Peaceman-Rachford splitting algorithm to the fixed point. However, approaches of this type require restrictions on weights that can only be enforced with complex numerical techniques.
 
To avoid the above difficulties, we use in this study nonlinear Perron-Frobenius theory \cite{krause2015positive,lemmens2012nonlinear} as an alternative to traditional theory in Hilbert spaces. More specifically, we consider deep equilibrium layers that belong to the class of standard interference mappings. These mappings, introduced in the next section, have been widely used in wireless networks \cite{yates1995framework,schubert2011interference,stanczak2009fundamentals,you2020note,miretti2023ul,miretti2023fixed,shindoh2020some,shindoh2019structures,cavalcante2016elementary,cavalcante2019connections}. They are not necessarily nonexpansive in Hilbert spaces, but they are contractive (though not necessarily Lipschitz contractions) in some metric spaces defined on cones, such as the cone of positive vectors. Before introducing SI mappings and their applications in DEQ models, we first need to establish the notation and formally introduce the concept of DEQ layers.

\vspace{-0.7\baselineskip}
\section{Preliminaries}
The nonnegative cone and its interior (i.e., the positive cone) are denoted as, respectively, 
\begin{equation*}
    \sgens{n}_+:=\{ (x_1,\dots,x_n) \in \sgens{n} \; | \; (\forall k\in\{1,\ldots, n\}) \; x_k \geq 0\}.
\end{equation*} 
and
\begin{multline*}
   \op{int}(\sgens{n}_+):= \{ (x_1,\dots,x_n) \in \sgens{n}_+ \; |\\ \; (\forall k\in\{1,\ldots, n\}) \; x_k > 0\}.
\end{multline*}
Let $x,y \in \sgens{n}_+$. The partial ordering induced by the nonnegative cone is denoted as $x \leq y \Leftrightarrow y - x \in \sgens{n}_+$. In a similar way, for $x \neq y$, $x < y \Leftrightarrow y - x \in \sgens{n}_+$, and $x \ll y \Leftrightarrow y - x \in \op{int}(\sgens{n}_+)$. The fixed point set of a function $f: X \rightarrow Y$ with $Y$ and $X$ being subsets of a given set $S$ is denoted as
\begin{equation*}
    \text{Fix}(f) = \{x^\star \in X \; | \; f(x^\star) = x^\star \}.
\end{equation*}



\subsection{Deep Equilibrium Layers}
We now have all the necessary notation to introduce generic DEQ models.
\begin{definition} \label{deq_layer}
	A DEQ layer maps an input $x\in X \subset \sgens{n}$ to an output $z^\star \in \textup{Fix}(g_x) \subset Y \subset X$, where $g_x: X\to Y$ is an explicit function given by
	\begin{equation}\label{deq_eq}
		g_x: X\to Y:z \mapsto \sigma(Wz + x);
	\end{equation}
	$W\colon X\to Y$ is a linear operator (weight matrix); and $\sigma\colon X\to Y$ is a (vector-valued nonlinear) activation function, composed elementwise from a given scalar activation function. 
\end{definition}

In the above definition, closed-form expressions for the implicit function $x\mapsto z^\star \in \textup{Fix}(g_x)$ are not required, and the numerical scheme used to compute the output $z^\star \in \textup{Fix}(g_x)$ from a given input $x$ is not specified. As a result, to ensure that the implicit function $x\mapsto z^\star \in \textup{Fix}(g_x)$ is well-defined and independent of the numerical scheme used to compute the output, we require $\textup{Fix}(g_x)$ to be a singleton for every $x\subset X$. This fixed point formulation allows for direct implicit differentiation, which is crucial for training DEQ models \cite{bai2019deep}. In the text that follows, for convenience, we always refer to a DEQ layer using its explicit function $g_x$ because, with the above restriction of $\mathrm{Fix}(g_x)$ being a singleton, the implicit function $x\mapsto z^\star\in\mathrm{Fix}(g_x)$ is well-defined.

One of the simplest numerical schemes for computing fixed points, which is the numerical scheme we consider in this study, is the standard fixed point iteration, which, using the notation in Definition~\ref{deq_layer}, we can write as
\begin{equation}\label{deq_fixed}
	(\forall k \in \mathbb{N}) \; z_{k+1} = g_x(z_k) \; \text{with} \; z_1 \in X.
\end{equation}


\subsection{Standard Interference and Positive Concave Mappings}

The DEQ layers that we propose in this study are a proper subclass of standard interference mappings, defined as follows.

\begin{definition}\label{def:mappings}
	A mapping $g\colon\sgens{n}_+\to\text{int}(\sgens{n}_+)$ is said to be a standard interference if it is
	\begin{enumerate}
		\item monotonic 
		\begin{equation}
			(\forall x\in\sgens{n}_+) (\forall\tilde{x}\in\sgens{n}_+)~ x\leq\tilde{x}\implies g(x)\leq g(\tilde{x}),\text{ and}
		\end{equation}
		\item scalable
		\begin{equation}
			(\forall x\in\sgens{n}_+) (\forall \lambda> 1) \quad g(\lambda x) \ll \lambda g(x).
		\end{equation}
	\end{enumerate}
\end{definition}

\begin{remark}
	Monotonic mappings in the sense of Definition \ref{def:mappings} are also known as order-preserving mappings in the mathematical literature, and they should not be confused with monotone operators used in convex analysis \cite{ryu2016primer,bauschke2017correction}, which is a different concept in general.
\end{remark}

SI mappings have at most one fixed point \cite{yates1995framework}, and its existence can be established with Proposition \ref{existence} in the appendix, which uses the concepts of asymptotic mappings (Definition \ref{asymptotic_mapping}) and nonlinear spectral radius (Definition \ref{nonlinear_sr}). A proper subclass of SI mappings is the class of the positive concave mappings (see Proposition \ref{pc_a0123}), defined below.





\begin{definition}\label{def_pc}
	Let $g: \sgens{n}_+\to\sgens{n}_+$ be concave with respect to the cone order; i.e.,
	\begin{equation}
		\begin{split}
			(\forall x\in\sgens{n}_+) (\forall{y}\in\sgens{n}_+)(\forall t \in (0,1)) & \\
			g(tx + (1-t)y) & \geq  tg(x) +  (1-t)  g(y). 
		\end{split}
	\end{equation}
	Then $g$ is called a nonnegative concave (NC) mapping. Furthermore, if the codomain of $g$ is in the set of positive vectors (i.e., $g: \sgens{n}_+\to\textup{int}(\sgens{n}_+)$), then $g$ is called a positive concave (PC) mapping. 
\end{definition}
We emphasize that NC mappings are not SI mappings in general (e.g., $x\mapsto x$ for $x\in\sgens{n}_+$ is an NC, but not an SI mapping), only PC mappings are guaranteed to be SI mappings without any further assumptions, see \cite{cavalcante2016elementary,cavalcante2019connections}.

An important property of PC mappings, which we exploit in this study, is that, if a fixed point exists, the fixed point iteration is guaranteed to converge geometrically fast to the fixed point (Proposition \ref{pc_geom}).

\section{PC Deep Equilibrium Layers} \label{pc}
We now proceed to construct the proposed pcDEQ layers $g_x$ based on the theory described in the previous section. In particular, recall that, by restricting DEQ layers to the class of PC mappings, we obtain simple conditions to guarantee the existence and uniqueness of the fixed point $z^\star$, and we also have the simple iterative scheme in (\ref{deq_fixed}) to compute the fixed point, which is an algorithm that converges geometrically fast for PC mappings. We start by restricting the activation functions to be nonnegative concave in the nonnegative cone, and, for later reference, the next remark gives examples of such functions.  


\begin{remark} \label{activ}
	We divide the activation functions allowed by the proposed framework into two classes: nonnegative concave functions and positive concave activation functions. Common examples in the neural network literature are shown in the two lists below.
	\begin{enumerate}
		\item continuous NC activation functions ($\sgens{}_+\to\sgens{}_+$):
		\begin{itemize}
			\item (ReLU6)
			$x\mapsto \min\{x,6\}$
			\item (hyperbolic tangent)
			$x\mapsto tanh \;{x}$
			\item (softsign)
			$x\mapsto \frac{x}{1+x}$
		\end{itemize}  
		\item continuous PC activation functions ($\sgens{}_+\to \text{int}(\sgens{}_+)$):
		\begin{itemize}
			\item (sigmoid)
			$x\mapsto \frac{1}{1+\exp{(-x)}}$  
		\end{itemize}
	\end{enumerate}
\end{remark}

We recall that the activation functions from the above lists are applied elementwise to vector arguments, see Definition~\ref{deq_layer}. The following lemma provides two simple sufficient conditions to construct positive concave DEQ layers.

\begin{lemma}\label{deq_pc}
Consider a DEQ layer $g_x\colon\sgens{n}_+\to\op{int}(\sgens{n}_+)$ of the form in (\ref{deq_eq}) in Definition~\ref{deq_layer} for an input $x$. Then:
\begin{itemize}
	\item (Assumption 1) $z \mapsto g_x(z) := \sigma(Wz + x)$ is a PC mapping $g_x\colon\sgens{n}_+\to\op{int}(\sgens{n}_+)$ if $z\in\sgens{n}_+$, $W \in \sgens{n \times n}_+$, $x \in \textup{int}(\sgens{n}_+)$, and $\sigma$ is constructed elementwise from any scalar activation function from List 1 in Remark~\ref{activ};
	\item (Assumption 2)  $z \mapsto g_x(z) := \sigma(Wz + x)$ is a PC mapping $g_x\colon\sgens{n}_+\to\op{int}(\sgens{n}_+)$ if $z\in\sgens{n}_+$, $W \in \sgens{n \times n}_+$, $x \in \sgens{n}_+$, and $\sigma$ is constructed elementwise from the scalar activation function from List 2 in Remark \ref{activ}.
\end{itemize}
\end{lemma}
\proof Let $f_x\colon\sgens{n}_+\to\sgens{n}_+$ be given by $z\mapsto Wz+x$ and let $\sigma\colon\sgens{n}_+\to\sgens{n}_+$ be the activation function. Let $f_x$ and $\sigma$ satisfy either Assumption 1 or Assumption 2, so that $g_x=\sigma\circ f_x.$ We note that both $f_x$ and $\sigma$ belong to the class of NC mappings. Let $z_1,z_2\in\sgens{n}_+$ and $t\in(0,1).$ Then, by concavity of $f_x$:
\begin{equation*}
	f_x(tz_1 + (1-t)z_2) \geq tf_x(z_1) + (1-t)f_x(z_2).
\end{equation*}
Hence, monotonicity of $\sigma$ implies that
\begin{equation*}
	\sigma[f_x(tz_1 + (1-t)z_2)] \geq \sigma[tf_x(z_1) + (1-t)f_x(z_2)].
\end{equation*}
From the concavity of $\sigma$ we deduce
\begin{equation*}
	\sigma[tf_x(z_1) + (1-t)f_x(z_2)]\geq t\sigma(f_x(z_1))+(1-t)\sigma(f_x(z_2)).
\end{equation*}
By combining the above two inequalities, we conclude that
\begin{equation*}
	(\sigma\circ f_x)(tz_1 + (1-t)z_2)\geq t(\sigma\circ f_x)(z_1)+(1-t)(\sigma\circ f_x)(z_2),
\end{equation*}
implying that $g_x=\sigma\circ f_x$ is an NC mapping under either Assumption 1 or Assumption 2. To show that $g_x$ is a PC mapping, we first note that, by the monotonicity of $f_x$, one has $f_x(z)\geq f_x(0)$ for $z\in\sgens{n}_+.$

Under Assumption 1, we have $y_0:=f_x(0)=x\gg 0$, so that $(\forall z\in\sgens{n}_+)\ f_x(z)\gg 0.$ We now note that a selection of any of the scalar activation functions from List 1 for $\sigma$ satisfies $(\forall v\gg 0)\ \sigma(v)\gg 0$, so that $\sigma(y_0)=(\sigma\circ f_x)(0)=g_x(0)\gg 0.$ Since $g_x$ is monotonic, one has $(\forall y\in\sgens{n}_+)\ g_x(y)\geq g_x(0)\gg 0$, thus $g_x$ is a PC mapping.

On the other hand, if Assumption 2 is satisfied, we note that, in this case, $\sigma(0)\gg 0$, hence $(\forall z\in\sgens{n}_+)\ g_x(z)=(\sigma\circ f_x)(z)\gg 0$, which implies $g_x$ is also a PC mapping in this case. $\ \blacksquare$

For convenience, the DEQ layers satisfying the assumptions in Lemma \ref{deq_pc} are formally defined below.

\begin{definition} \label{pcDEQ_def}
Let $g_x\colon\sgens{n}_+\to\textup{int}(\sgens{n}_+)$ be a DEQ layer of the form in (\ref{deq_eq}) for a given input $x$. If $g_x$ (including the input $x$) satisfies Assumption 1 in Lemma \ref{deq_pc}, then $g_x$ is called a pcDEQ-1 layer. If $g_x$ satisfies Assumption~2 in Lemma \ref{deq_pc}, then $g_x$ is called a pcDEQ-2 layer. By pcDEQ we mean a layer $g_x$ that is either a pcDEQ-1 layer or a pcDEQ-2 layer.
\end{definition}

\begin{figure}[h!]
	    \centering
	     \begin{subfigure}
		         \centering
		         \includegraphics[scale=0.85]{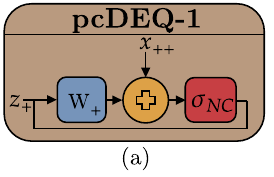}
		     \end{subfigure}
	     \begin{subfigure}
		         \centering
		         \includegraphics[scale=0.85]{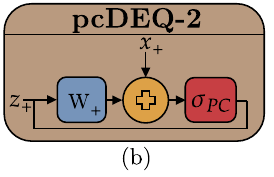}
		     \end{subfigure}
	    \caption{The visualization of the possible construction of pcDEQ layers. The symbols shown in the figures mean: $\sgens{n \times n}_+\ni W_+$ - nonnegative weights, $\sgens{n}_+\ni z_+$ - nonnegative vector of fixed point iteration, $\sgens{n}_+\ni x_{+}$ - nonnegative input to the layer, $\textup{int}(\sgens{n}_+)\ni x_{++}$ - positive input to the layer, $\sigma_{NC}$ - nonnegative concave activation function (List 1 in Remark \ref{activ}) and $\sigma_{PC}$ - positive concave activation function (List 2 in Remark \ref{activ}).}
	    \label{fig:pcDEQ}
	\end{figure}

PcDEQ-1 and pcDEQ-2 layers are illustrated in Figure \ref{fig:pcDEQ}.

Lemma \ref{deq_pc} asserts that pcDEQ layers are PC mappings for any allowed input $x$, which, in view of Proposition \ref{pc_a0123}, implies that they are also SI mappings. We can now use Definition \ref{asymptotic_mapping} to provide the form of the asymptotic mapping associated with a pcDEQ layer. The assertion of Proposition \ref{asym_value} follows from Proposition~11 in \cite{piotrowski2024fixed}, and, for completeness, we include the proof below.

\begin{proposition} \label{asym_value}
For a given input $x$, let $g_{x,\infty}\colon\sgens{n}_+\to\sgens{n}_+$ be the asymptotic mapping (in the sense of Definition \ref{asymptotic_mapping}) of a pcDEQ layer $g_x\colon\sgens{n}_+\to\op{int}(\sgens{n}_+)$ satisfying Definition \ref{pcDEQ_def}. Then $g_{x,\infty}(z) = 0$ for every $z\in\sgens{n}_+$.
\end{proposition}
\proof The asymptotic mapping $g_{x,\infty}$ according to Definition \ref{asymptotic_mapping} is defined as follows
\begin{equation}
	g_{x,\infty}(z) = \lim_{p\to\infty} \frac{1}{p}g_x(p z) = \lim_{p\to\infty} \frac{\sigma(Wp z + x)}{p}.
\end{equation}
Applying L'Hôpital's rule to the activation function $\sigma$ with $p$ as an argument, we have
\begin{equation}
	g_{x,\infty}(z) = \lim_{p\to\infty}\sigma'(Wpz + x)Wz.
\end{equation}
For activation functions composed from any scalar activation functions in Remark \ref{activ}, we have $\lim_{p\to\infty} \sigma'(Wpz + x) = 0$, hence also
\begin{equation}
	\lim_{p\to\infty}\sigma'(Wpz + x)Wz = 0.\ \blacksquare
\end{equation}


The following Corollary \ref{sr_corr} follows directly from Proposition \ref{asym_value}.

\begin{corollary}\label{sr_corr}
With notation in Proposition \ref{asym_value}, let $\rho(g_{x,\infty})\in\sgens{}_+$ be the spectral radius of $g_{x,\infty}$ in the sense of Definition \ref{nonlinear_sr} for a given input $x$. Then $\rho(g_{x,\infty}) = 0$.
\end{corollary}

Based on the previous results, Proposition \ref{exist_uniq_deq} establishes the existence and uniqueness of the fixed point for the pcDEQ layers, in addition to the geometric convergence of the fixed point iteration.

\begin{proposition}\label{exist_uniq_deq}
Let $g_x\colon\sgens{n}_+\to\op{int}(\sgens{n}_+)$ be a pcDEQ layer satisfying Definition \ref{pcDEQ_def}. Then $g_x$ has a unique fixed point $z^\star$ for every input $x$ satisfying the conditions in Definition \ref{pcDEQ_def}.
Moreover, the fixed point iteration of $g_x$ in (\ref{deq_fixed}) converges geometrically in the sense of Definition \ref{geom} to $z^\star$ for any $z_1 \in \sgens{n}_+$ and any input $x$.  
\end{proposition}
\proof Choose any input $x$ satisfying the conditions in Definition \ref{pcDEQ_def}. From Lemma \ref{deq_pc}, $g_x$ is a PC mapping. Thus, from Proposition \ref{pc_a0123}, $g_x$ is also an SI mapping. From Corollary~\ref{sr_corr}, the nonlinear spectral radius of $g_x$ is $\rho(g_{x,\infty}) = 0.$ If follows from Proposition \ref{existence} that $g_x$ has a unique fixed point $z^\star.$ Then, from Proposition \ref{pc_geom} it follows that the fixed point iteration of $g_x$ in (\ref{deq_fixed}) converges geometrically to $z^\star$ with a factor $c \in [0,1)$ for any starting point $z_1 \in \sgens{n}_+.$ $\ \blacksquare$

\section{Experiments}\label{experiments}
The experiments were carried out on three commonly known computer vision datasets: MNIST, SVHN, and CIFAR-10. The proposed pcDEQ\footnote{The \textit{Pytorch} source code of pcDEQs along with examples of use is available at the following link: \url{https://github.com/mateuszgabor/pcdeq}} models were compared with competitive approaches for which the uniqueness of the fixed point is mathematically established, namely: monotone operator deep equilibrium models (monDEQ) \cite{winston2020monotone}, neural ordinary differential equations (NODE) \cite{chen2018neural}, and augmented neural differential equations (ANODE) \cite{dupont2019augmented}. The comparison was performed according to the results reported in \cite{dupont2019augmented,winston2020monotone}. For pcDEQ models, the fixed point $z^\star$ is computed using the standard fixed point iteration in (\ref{deq_fixed}). The stopping criterion of fixed point iteration is based on the relative error, calculated as $\frac{||z_{k+1} - z_k||}{||z_{k+1}||} \leq \epsilon$, where $||\cdot||$ is a Frobenius norm and $\epsilon$ is a tolerance. In our experiments, the tolerance $\epsilon$ was set to $1e-4$.

The experiments were performed using the Google Colab platform with a NVIDIA Tesla T4 16GB GPU.

\subsection{Architecture Setup}
According to Definition \ref{pcDEQ_def}, we consider two options to build pcDEQ layers, pcDEQ-1 (Figure \ref{fig:pcDEQ} (a)) and pcDEQ-2 (Figure \ref{fig:pcDEQ} (b)). For both options, the nonnegative weights $W_+$ are achieved by projecting the negative values to zeros after backpropagation. To construct pcDEQ-1, it is necessary to provide a positive input $x_{++}$ (see Figure \ref{fig:pcDEQ} (a)), which is performed by applying elementwise the softplus activation function ($\sigma\colon\sgens{}\to\text{int}(\sgens{}_+)$) before the pcDEQ-1 layer, as the activation functions in networks with pcDEQ-1 layers are functions from List 1 in Remark \ref{activ}.
In the case of the pcDEQ-2 model, the activation function is a PC activation function from List 2 in Remark \ref{activ}. For pcDEQ-2, the nonnegativity of $x_+$ is achieved by applying the ReLU activation function before the DEQ layer. 

Compared to monDEQ models, the construction of pcDEQ layers is much simpler and does not require a special layer implementation. MonDEQ layer has to be carefully parameterized by a set of two weights to satisfy the assumptions of strong monotonicity, which results in overparameterization and a more complicated implementation. Furthermore, computing convolutions in monDEQ models requires calculating fast Fourier transforms, which, as noted in \cite{winston2020monotone}, are empirically 2-3 times slower than computing convolutions in a standard manner. On the other hand, in pcDEQ models, as mentioned previously, the only requirement is to use concave activation functions and constrain the weights to be nonnegative. Such a parameterization does not produce extra computational overhead compared to standard DEQ models.

In experiments, we consider three types of pcDEQ models with four activation functions from Remark \ref{activ}. The first type of network is based on using one linear pcDEQ layer. The second type of network is based on using one convolutional pcDEQ layer, and the last type is based on the use of three convolutional pcDEQ layers, between which the explicit downsampling layers occur. The architectural details are discussed in Appendix \ref{arch_details}, and the experimental hyperparameters for each network in Appendix \ref{hyper}. Networks using a single linear pcDEQ layer have “L” suffix in their name. A similar scenario is for networks with a single convolutional pcDEQ layer, to network name suffix “SC” is added. For networks with three convolutional pcDEQ layers, the suffix “MC” is added to the network name. The suffix “MT” in monDEQs refers to the multi-tier architecture used in \cite{winston2020monotone}.

\subsection{Results}
Tables \ref{tab:mnist}, \ref{tab:svhn}, and \ref{tab:cifar10} show the results obtained by pcDEQ models with compared methods for MNIST, SVHN and CIFAR-10, respectively. As the results show, the architectures based on pcDEQ achieve competitive results compared to other implicit models. In each scenario, pcDEQ models were constructed with a smaller number of parameters compared to NODE, ANODE, and monDEQ. In the case of the results obtained on the MNIST dataset (Table \ref{tab:mnist}), all pcDEQ configurations outperform the NODE, ANODE, and monDEQ approaches. For the results obtained on the SVHN dataset (Table \ref{tab:svhn}), it can be seen that the highest accuracy is obtained by pcDEQ with ReLU6 activation functions and three convolutional layers.
In the case of results obtained on the CIFAR-10 dataset, Table \ref{tab:cifar10} shows that with a lower number of parameters, pcDEQ models can achieve similar or better results compared to NODE, ANODE, or monDEQ. Moreover, similar to previous works, we trained larger pcDEQ models with data augmentation on CIFAR-10. As can be seen, for this setup, pcDEQ with three pcDEQ convolutional layers and softsign activation functions achieves the highest accuracy among all pcDEQ configurations.

\begin{table}[h!]
	\centering
	\caption{Test accuracies of pcDEQ models averaged over five runs on MNIST dataset compared with NODE, ANODE and monDEQ; $\dagger$ as reported in \cite{dupont2019augmented}; $\ddagger$ as reported in \cite{winston2020monotone}.}
	\label{tab:mnist}
	\begin{tabular}{@{}lcc@{}}
		\toprule
		\multirow{2}{*}{\textbf{Method}} & \multicolumn{2}{c}{\textbf{MNIST}} \\ \cmidrule(l){2-3} 
		& \#Parameters & Accuracy {[}\%{]} \\ \midrule
		NODE$^\dagger$  & 84K & 96.4\\
		ANODE$^\dagger$  & 84K & 98.2  \\
		monDEQ-L$^\ddagger$  & 84K & 98.1  \\
		monDEQ-SC$^\ddagger$  & 84K & 99.1 \\
		monDEQ-MT$^\ddagger$  & 81K & 99.0 \\
		pcDEQ-1-L-ReLU6 & 70K & 98.1   \\
		pcDEQ-1-L-Tanh & 70K & 98.2  \\
		pcDEQ-1-L-Softsign & 70K & 98.1 \\
		pcDEQ-2-L-Sigmoid & 70K & 98.1 \\
		pcDEQ-1-SC-ReLU6 & 69K & 99.2 \\
		pcDEQ-1-SC-Tanh & 69K & 99.2 \\
		pcDEQ-1-SC-Softsign & 69K & 99.1 \\
		pcDEQ-2-SC-Sigmoid & 69K & 98.9 \\
		pcDEQ-1-MC-ReLU6 & 41K & 99.3 \\
		pcDEQ-1-MC-Tanh & 41K & 99.2 \\
		pcDEQ-1-MC-Softsign & 41K & 99.2 \\
		pcDEQ-2-MC-Sigmoid & 41K & 98.7 \\ \bottomrule
	\end{tabular}
\end{table}

\begin{table}[h!]
	\centering
	\caption{Test accuracies of pcDEQ models averaged over five runs on SVHN dataset compared with NODE, ANODE and monDEQ; $\dagger$ as reported in \cite{dupont2019augmented}; $\ddagger$ as reported in \cite{winston2020monotone}.}
	\label{tab:svhn}
	\begin{tabular}{@{}lcc@{}}
		\toprule
		\multirow{2}{*}{\textbf{Method}} & \multicolumn{2}{c}{\textbf{SVHN}} \\ \cmidrule(l){2-3} 
		& \#Parameters & Accuracy {[}\%{]} \\ \midrule
		NODE$^\dagger$  & 172K & 81.0 \\
		ANODE$^\dagger$  & 172K & 83.5 \\
		monDEQ-SC$^\ddagger$  & 172K & 88.7 \\
		monDEQ-MT$^\ddagger$  & 170K & 92.4 \\
		pcDEQ-1-SC-ReLU6 & 165K & 88.0 \\
		pcDEQ-1-SC-Tanh & 165K & 88.1 \\
		pcDEQ-1-SC-Softsign & 165K & 88.4 \\
		pcDEQ-2-SC-Sigmoid & 165K & 87.3 \\
		pcDEQ-1-MC-ReLU6 & 131K & 93.0 \\
		pcDEQ-1-MC-Tanh & 131K & 92.3 \\
		pcDEQ-1-MC-Softsign & 131K & 92.3 \\
		pcDEQ-2-MC-Sigmoid & 131K & 91.5 \\ \bottomrule
	\end{tabular}
\end{table}

\begin{table}[h!]
	\centering
	\caption{Test accuracies of pcDEQ models averaged over five runs on CIFAR-10 dataset compared with NODE, ANODE and monDEQ; $\dagger$ as reported in \cite{dupont2019augmented}; $\ddagger$ as reported in \cite{winston2020monotone}; * with data augmentation. }
	\label{tab:cifar10}
	\begin{tabular}{@{}lcc@{}}
		\toprule
		\multirow{2}{*}{\textbf{Method}} & \multicolumn{2}{c}{\textbf{CIFAR-10}} \\ \cmidrule(l){2-3} 
		& \#Parameters & Accuracy {[}\%{]} \\ \midrule
		NODE$^\dagger$  & 172K & 53.7 \\
		NODE$^\ddagger$* & 1M & 59.9 \\
		ANODE$^\dagger$  & 172K & 60.6 \\
		ANODE$^\ddagger$* & 1M & 73.4 \\
		monDEQ-SC$^\ddagger$  & 172K & 74.0 \\
		monDEQ-SC$^\ddagger$* & 854K & 82.0 \\
		monDEQ-MT$^\ddagger$ & 170K & 72.0 \\
		monDEQ-MT$^\ddagger$* & 1M & 89.0 \\
		pcDEQ-1-SC-ReLU6 & 165K & 76.3 \\
		pcDEQ-1-SC-Tanh & 165K & 76.6 \\
		pcDEQ-1-SC-Softsign & 165K & 76.4 \\
		pcDEQ-2-SC-Sigmoid & 165K & 75.5 \\
		pcDEQ-1-MC-ReLU6 & 131K & 78.2 \\
		pcDEQ-1-MC-ReLU6* & 661K & 89.2 \\
		pcDEQ-1-MC-Tanh & 131K & 76.5 \\
		pcDEQ-1-MC-Tanh* & 661K & 88.5 \\
		pcDEQ-1-MC-Softsign & 131K & 77.1 \\
		pcDEQ-1-MC-Softsign* & 661K & 89.0 \\
		pcDEQ-2-MC-Sigmoid & 131K & 71.0 \\
		pcDEQ-2-MC-Sigmoid* & 661K & 85.6 \\ \bottomrule
	\end{tabular}
\end{table}

\begin{figure*}[h!]
	     \centering
	     \begin{subfigure}
		         \centering
		         \includegraphics[scale=0.375]{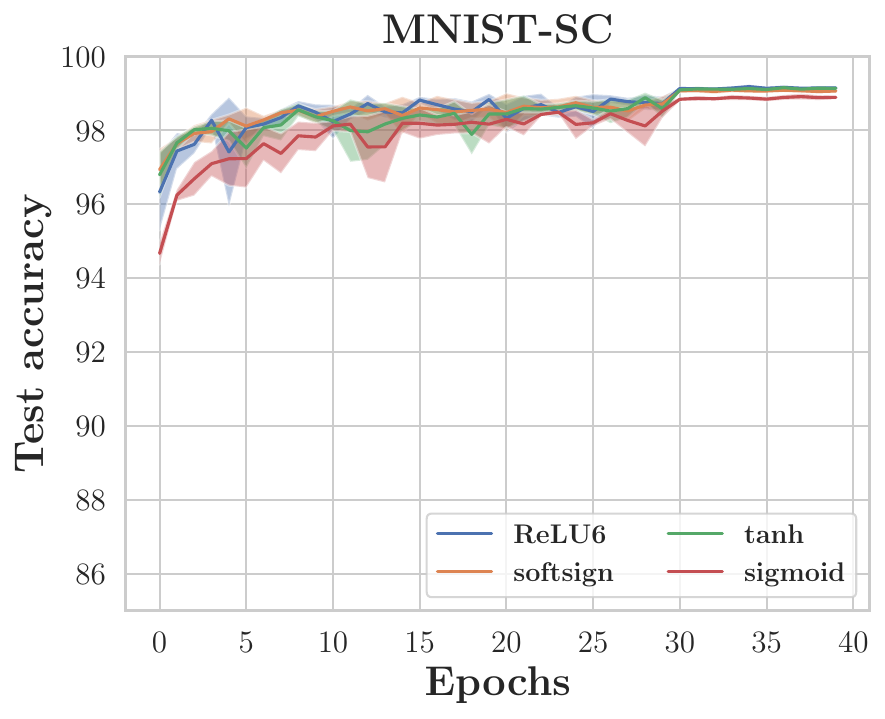}
		     \end{subfigure}
	     \hfill
	     \begin{subfigure}
		         \centering
		         \includegraphics[scale=0.375]{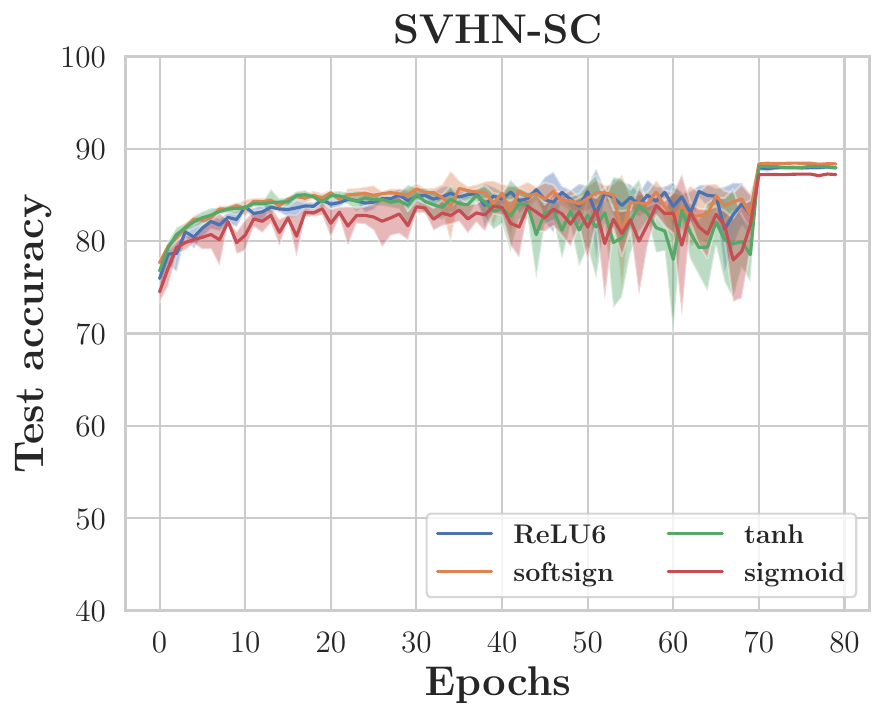}
		     \end{subfigure}
	     \hfill
	     \begin{subfigure}
		         \centering
		         \includegraphics[scale=0.375]{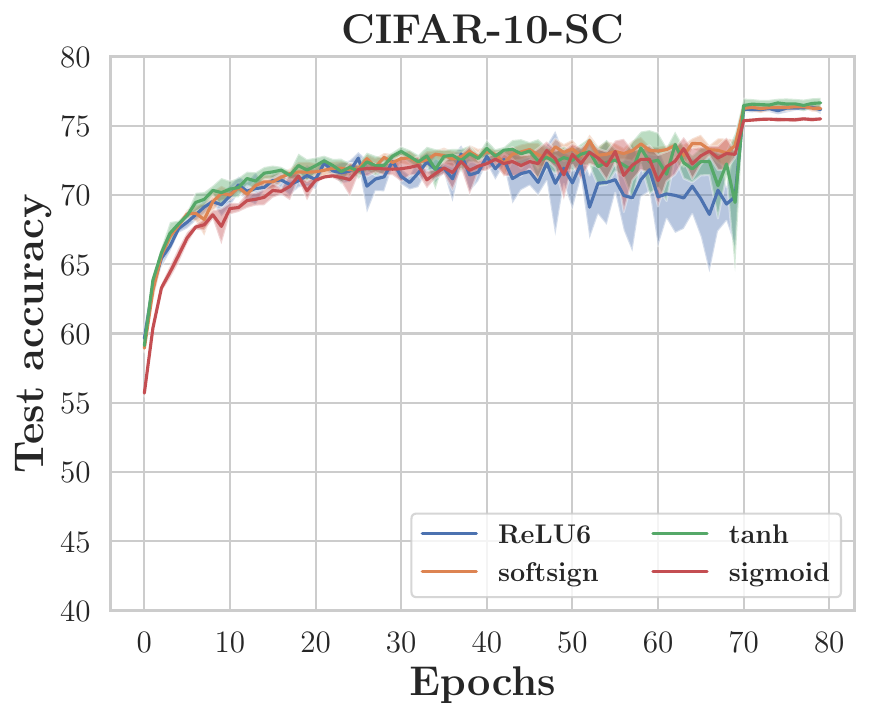}
		     \end{subfigure}
	     \caption{Test accuracies during training for the pcDEQ model with a single convolutional layer over five experiment runs.}
	     \label{sc_training_curves}
	\end{figure*}

Figure \ref{sc_training_curves} shows the training curves for the pcDEQ models with a single convolutional layer. For other cases, the figures are attached in Appendix \ref{add_results}.

\subsection{Convergence Analysis} \label{conv}
\begin{figure*}[h!]
	     \centering
	     \begin{subfigure}
		         \centering
		         \includegraphics[scale=0.375]{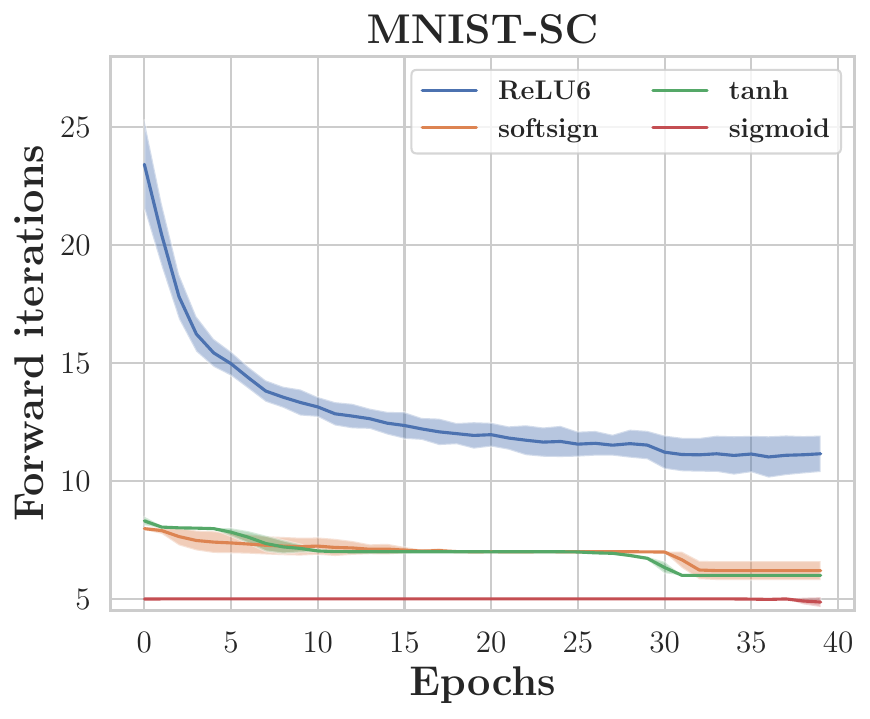}
		     \end{subfigure}
	     \hfill
	     \begin{subfigure}
		         \centering
		         \includegraphics[scale=0.375]{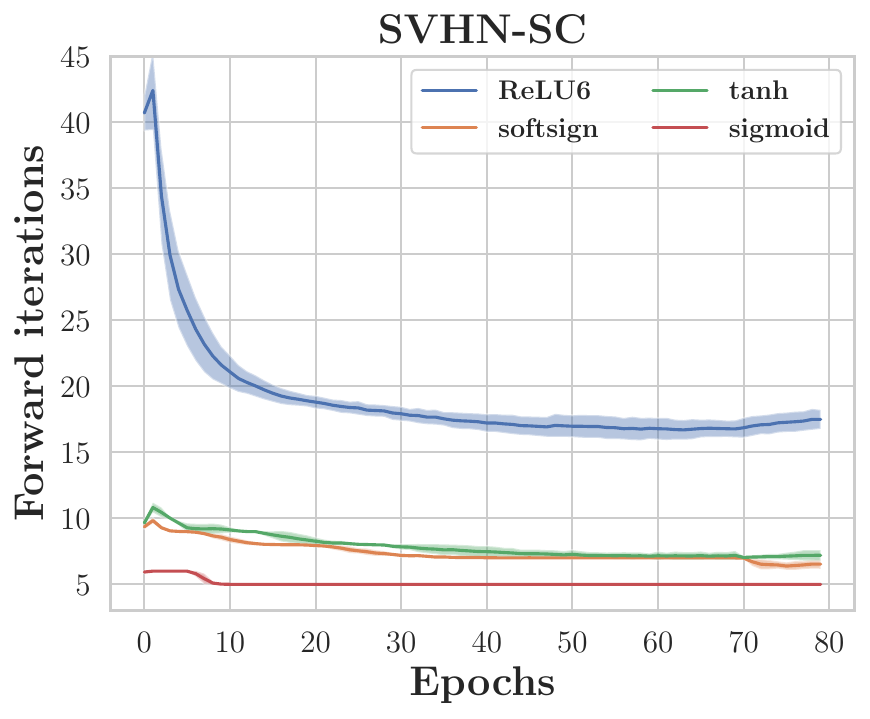}
		     \end{subfigure}
	     \hfill
	     \begin{subfigure}
		         \centering
		         \includegraphics[scale=0.375]{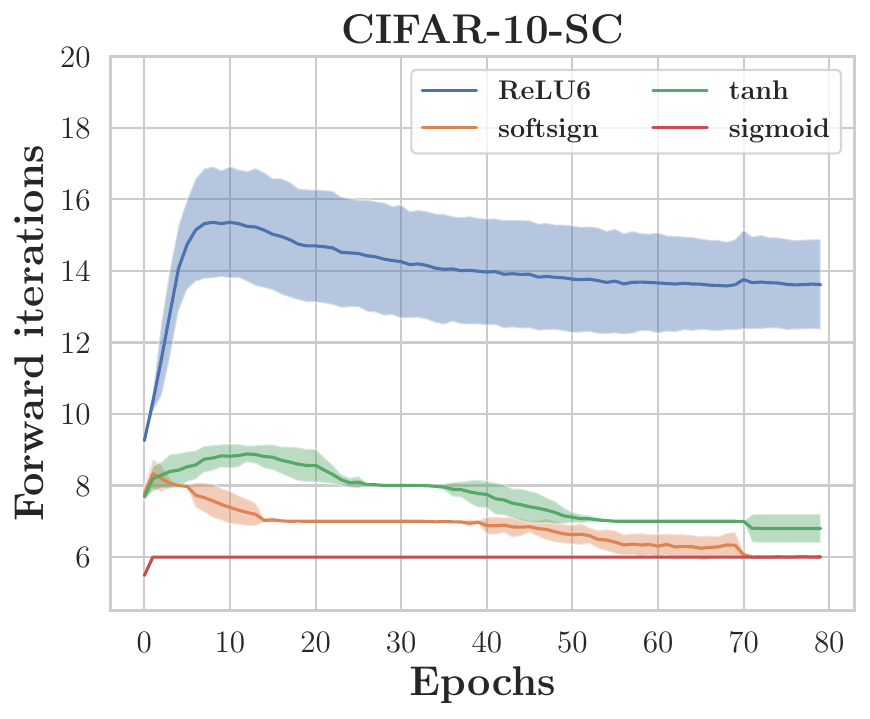}
		     \end{subfigure}
	
	     \begin{subfigure}
		         \centering
		         \includegraphics[scale=0.375]{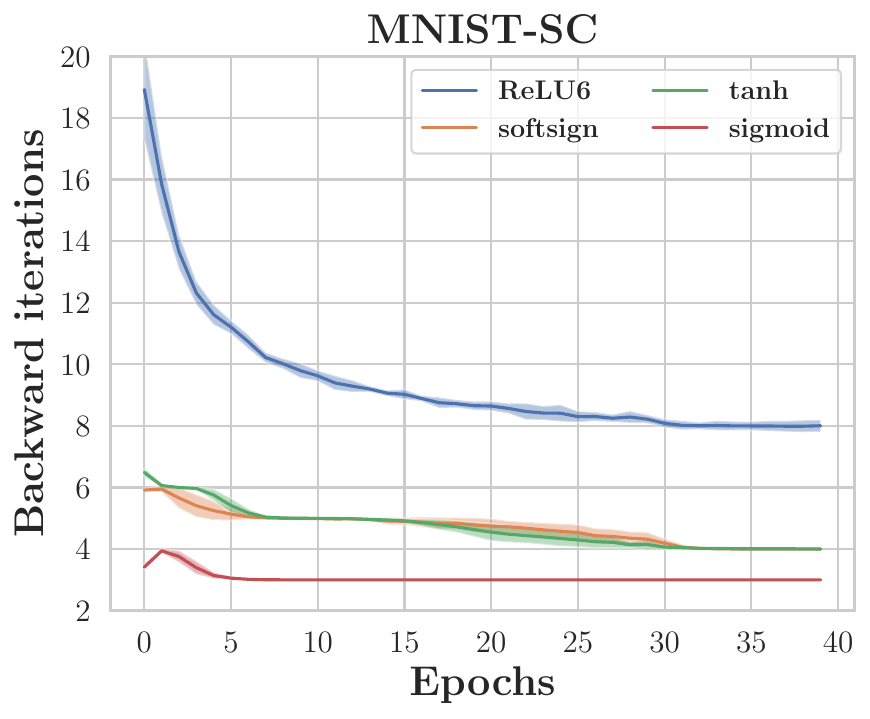}
		     \end{subfigure}
	     \hfill
	     \begin{subfigure}
		         \centering
		         \includegraphics[scale=0.375]{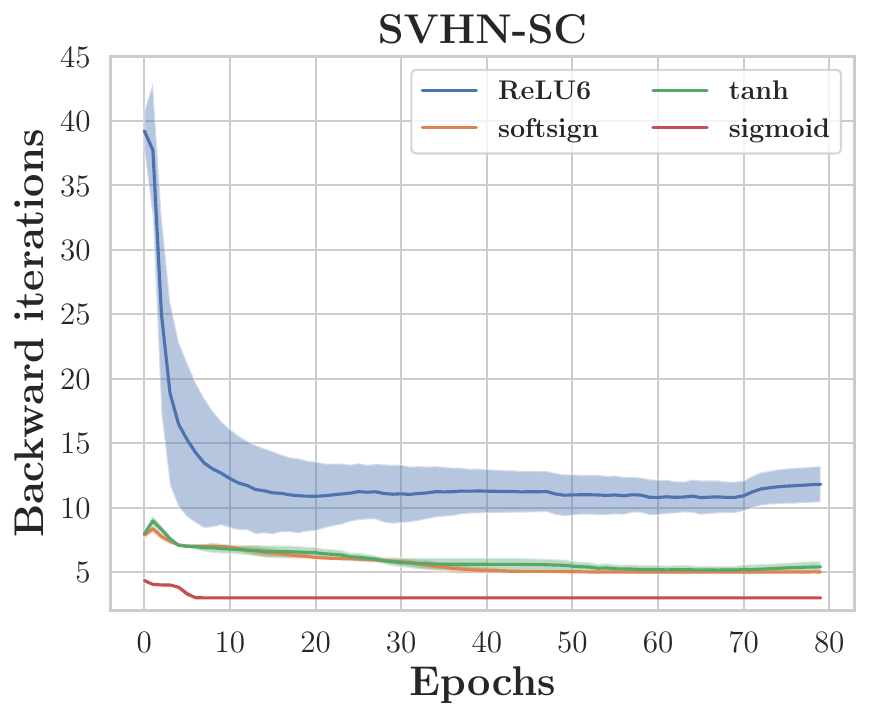}
		     \end{subfigure}
	     \hfill
	     \begin{subfigure}
		         \centering
		         \includegraphics[scale=0.375]{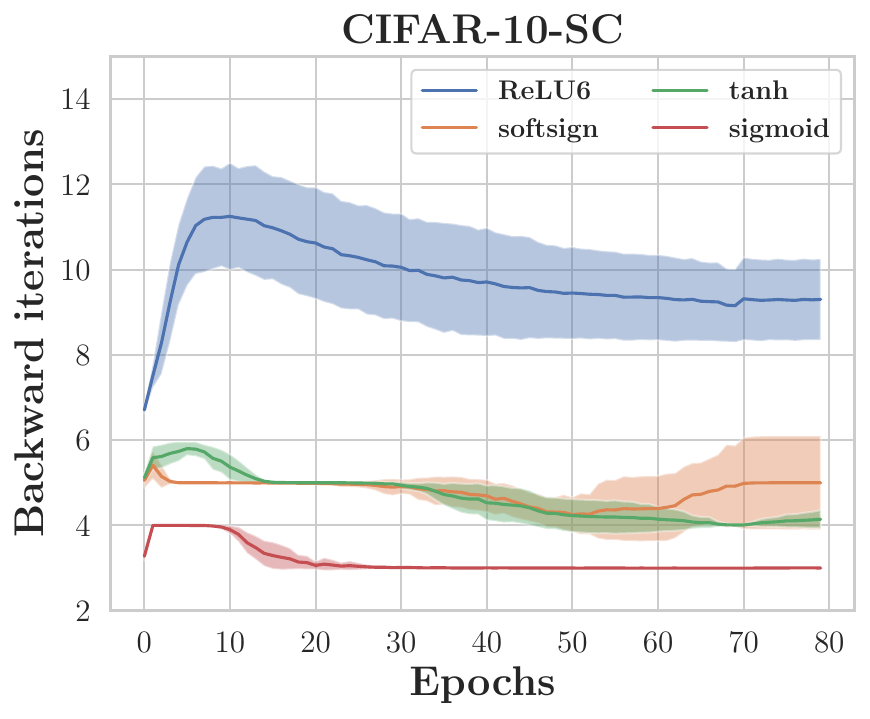}
		     \end{subfigure}
	
	     \caption{Number of fixed point iterations for computing the fixed point in forward and backward passes for the pcDEQ model with a single convolutional layer over five experiment runs.}
	     \label{fig:sc_iterations}
\end{figure*}

Figure \ref{fig:sc_iterations} shows the average number (among all batches in epoch) of fixed point iterations to compute a fixed point per epoch for the pcDEQ model with a single convolutional layer. From this figure, it can be seen that the convergence is very fast and accurate. For pcDEQ with sigmoid activation function, the fixed point iteration satisfies the stopping criterion with less than eight iterations. From Proposition \ref{exist_uniq_deq}, we know that the convergence of the fixed point iteration for the pcDEQ models is geometric. We could improve the convergence rate with vector extrapolation techniques. This fact is very interesting because such fast convergence is achieved without using any acceleration method. The interesting fact is that in standard DEQ and monDEQ models, the number of iterations of the used iterative method increases during training; in the proposed pcDEQ architectures, this effect does not occur. For other architecture configurations, the situation is similar to that described in this section, and the results are given in Appendix \ref{add_results}.

\subsection{Lipschitz Continuity}

In this section, by considering the standard Euclidean metric space, we show empirically that the assumptions used to determine the uniqueness of the fixed point are weaker compared to the standard assumptions in convex analysis. In general, to determine the uniqueness of the fixed point based on the Banach fixed point theorem, we need information about a Lipschitz constant of the function, and, in the machine learning literature, we often consider the standard Euclidean space. In the case of neural networks, if the used activation functions are nonexpansive (such as those in Remark \ref{activ}), then a Lipschitz constant $L$ w.r.t. the standard Euclidean space is upper bounded by the product of spectral norms of weight operators \cite{combettes2020lipschitz} as follows:
\begin{equation}\label{lipschitz}
	L \leq \prod_{i=1}^m||W_i||,
\end{equation}
where $m$ is the number of weight operators and $||\cdot||$ is the spectral norm.

\begin{figure}[h!]
	    \centering
	    \includegraphics[scale=0.5]{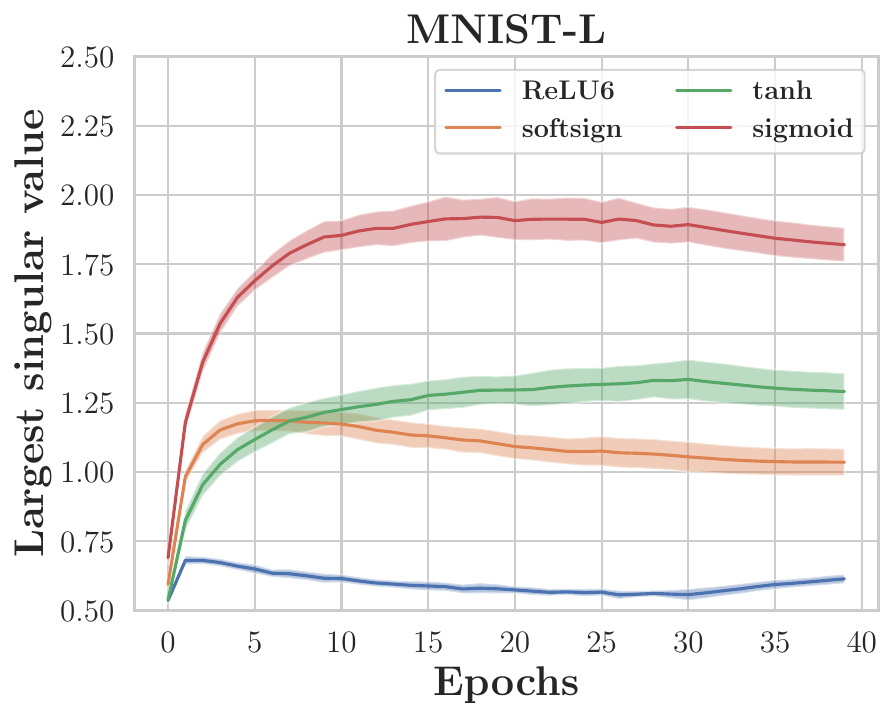}
	    \caption{Largest singular value of pcDEQ linear layer over five experiment runs.}
	    \label{fig:lip}
	\end{figure}

For the simple scenario, we investigated the pcDEQ model containing one linear pcDEQ layer ($m=1)$ with four activation functions. The average largest singular value per epoch for the entire training process is shown in Figure \ref{fig:lip}. It can be concluded that for the network with softsign, hyperbolic tangent, and sigmoid activation function, the Lipschitz constant in (\ref{lipschitz}) is $L > 1$ for most of the training time. For such a case, the uniqueness of the fixed point cannot be determined from the Banach fixed point theorem, because the Lipschitz constant $L$ in (\ref{lipschitz}) does not satisfy $L < 1$. In the case of a network with the ReLU6 activation function, the Lipschitz constant in (\ref{lipschitz}) satisfies $L < 1$ 
for the entire training process. In such a case, the Banach fixed point theorem can be used to determine the uniqueness of the fixed point and the linear convergence to it by using fixed point iteration. However, there are no guarantees that even for ReLU6 the Lipschitz constant $L$ will always be less than one, for example, with different learning rates, initialization, architecture, etc. On the other hand, the necessary and sufficient conditions of the uniqueness of the fixed point of SI mappings are weaker compared to the Banach fixed point theory, because the uniqueness of the fixed point is independent of a Lipschitz constant and the unique fixed point can still exist even if $L>1$.

\section{Limitations}
The proposed approach may appear restrictive due to the non-negativity constraint on weights and the use of concave activation functions. However, these restrictions are essential to ensure theoretical guarantees, such as the uniqueness of the fixed point and the convergence of the standard fixed point iteration. It is important to note that existing approaches offering similar guarantees, such as monDEQ models, also impose constraints to ensure that the mappings are strongly monotone in the sense used in convex analysis. The advantage of our approach is that the imposed restrictions are simpler to implement in practice compared to those based on monotone operator theory.


\section{Conclusions}
In this study, we have proposed a new class of DEQ models with guarantees of the existence of a unique fixed point. The parametrization of pcDEQ is very simple, and it does not require any sophisticated training modifications compared to standard DEQ models. Moreover, the fixed point can be easily computed with the standard fixed point iteration, and the convergence is guaranteed to be geometric. The proposed pcDEQ models are based on the theory of SI mappings, which are widely used in the wireless literature. To the best of our knowledge, this is the first practical application of SI mappings theory in the deep learning literature.

This study provides the foundation for extending the proposed method to a larger and less restricted class of mappings with guarantees similar to SI mappings. As an example, one may aim to weaken further the assumptions used to construct pcDEQ models to obtain more versatile models, capable of incorporating a wider class of weight operators and activation functions in DEQ layers. Such a direction seems feasible in view of the recent results on the existence and shape of the fixed point sets of (subhomogeneous) weakly standard interference (WSI) neural networks provided in \cite{piotrowski2024fixed}. Another promising future research in the direction proposed in this paper should focus on providing even stronger guarantees of convergence rate compared with the geometric convergence of the fixed point iteration of pcDEQ models established in this paper. Indeed, the empirical convergence analysis provided in Section \ref{conv} suggests that the actual rate of convergence may actually be linear.

\section*{Acknowledgments}
This work is supported by the National Center of Science under Preludium 22 project No. UMO-2023/49/N/ST6/02697 (Mateusz Gabor), National Centre for Research and Development of Poland (NCBR) under grant EIG CONCERT-JAPAN/04/2021, and by the Federal Ministry of Education and Research of Germany under grant 01DR21009 and the programme ``Souver\"an. Digital. Vernetzt.'' Joint project 6G-RIC, project identification numbers: 16KISK020K and 16KISK030.

\section*{Impact Statement}
This paper presents work whose goal is to advance the field of Machine Learning. There are many potential societal consequences of our work, none which we feel must be specifically highlighted here.

\nocite{langley00}

\bibliography{example_paper}
\bibliographystyle{icml2024}

\newpage
\appendix
\onecolumn
\section{Known Results}
\begin{definition}\label{geom}
	\cite{ortega2000iterative}[Chapter 9]
	Let $(x_k: k \in \mathbb{N})$ be a sequence in $\sgens{n}$, then $(x_k: k \in \mathbb{N})$ converges geometrically to $x^\star \in \sgens{n}$ with a rate $c \in [0,1)$ and a constant $\gamma > 0$ if
	\begin{equation}
		\forall k \in \mathbb{N} \quad ||x_{k+1} - x^\star|| \leq \gamma c^k,
	\end{equation}
	where $|| \cdot ||$ is a given norm.
\end{definition}


\begin{definition}\label{asymptotic_mapping}
	\cite{cavalcante2019connections,oshime1992perron} Let $f\colon \sgens{n}_+\to\text{int}(\sgens{n}_+)$ be an SI mapping in the sense of Definition~\ref{def:mappings}. The asymptotic mapping associated with~$f$ is the mapping defined by
	\begin{equation}
		f_\infty: \sgens{n}_+ \to\sgens{n}_+: z \mapsto \lim_{p\to\infty} \frac{1}{p}f(p z).
	\end{equation}
	We recall that the above limit always exists and that the resulting asymptotic mapping $f_\infty$ is positively homogeneous; i.e., $(\forall\alpha>0)$ $(\forall z\in\sgens{n}_+)$ $f_\infty(\alpha x)=\alpha f_\infty(z).$ 
\end{definition}

\begin{definition}\label{nonlinear_sr}
	The (nonlinear) spectral radius of an SI mapping is defined as the largest eigenvalue of the corresponding asymptotic mapping \cite{cavalcante2019connections,oshime1992perron}, and it is given by
	\begin{equation}
		\begin{split}
			p(f_\infty) := 
			\text{max}\{\lambda \in  \sgens{}_+\;|\; \exists z \in \sgens{n}_+\setminus\{0\} \; s.t. \; f_\infty(z) = \lambda z\} \in \sgens{}_+.
		\end{split}
	\end{equation}
\end{definition}

\begin{proposition}\label{existence}
	\cite{cavalcante2019connections} Let $f\colon\sgens{n}_+\to\textup{int}(\sgens{n}_+)$ be an SI mapping. Then $\text{Fix}(f) \neq \emptyset$ if and only if $p(f_\infty) < 1$. Furthermore, if a fixed point exists, then it is positive and unique.
\end{proposition}

\begin{corollary}\label{concave_mon}
	\cite{cavalcante2016elementary} If $f\colon\sgens{n}_+\to\sgens{n}_+$ is an NC mapping or $f\colon\sgens{n}_+\to\textup{int}(\sgens{n}_+)$ is a PC mapping in the sense of Definition \ref{def_pc}, then $f$ is monotonic.
\end{corollary}

\begin{proposition}\label{pc_a0123}
	\cite{cavalcante2016elementary,cavalcante2019connections} If $f: \sgens{n}_+\to\textup{int}(\sgens{n}_+)$ is a PC mapping, then $f$ is an SI mapping.
\end{proposition}

\begin{proposition}\label{pc_geom}
	\cite{piotrowski2022fixed} Let $f\colon\sgens{n}_+\to \textup{int}(\sgens{n}_+)$ be a PC mapping with a fixed point $x^\star \in \textup{int}(\sgens{n}_+)$. Then, for any $x_1 \in \sgens{n}_+$, the fixed point iteration of $f$ converges geometrically to $x^\star$ with a factor $c \in [0,1)$ w.r.t. any metric induced by a norm in $\sgens{n}$.
\end{proposition}

\section{Architecture Details}\label{arch_details}
As noted in Section \ref{experiments}, we consider three types of architectures of pcDEQ models. Based on the results of previous DEQ papers \cite{bai2019deep,bai2020multiscale}, using some regularization techniques can prevent overfitting and improve the final results. Therefore, similar to standard DEQ models \cite{bai2019deep} for each pcDEQ layer, we apply weight normalization \cite{salimans2016weight}. Similarly to the official tutorial on implicit models \cite{kolter2020deep}, we also used batch normalization before and after the DEQ layer, which allows improving the final results. In Figures \ref{fig:fc_archl}, \ref{fig:sc_archl}, \ref{fig:mc_archl}, the meaning of the blocks is as follows:
\begin{itemize}
	\item Linear — linear layer,
	\item BN — batch normalization layer,
	\item pcDEQ-1 — pcDEQ-1 layer in Definition \ref{pcDEQ_def},
	\item pcDEQ-2 — pcDEQ-2 layer in Definition \ref{pcDEQ_def},
	\item Conv2D — 2D convolutional layer,
	\item Conv2D, s=2 —  2D convolutional downsampling layer with stride equal to 2,
	\item BN + Softplus — batch normalization layer followed by a softplus activation function,
	\item BN + ReLU — batch normalization followed by ReLU activation functions,
	\item MaxPool + BN — max pooling layer followed by batch normalization,
	\item AvgPool — average pooling layer.
\end{itemize}


\subsection{Architecture with Single pcDEQ Linear Layer}
\begin{figure}[h!]
	    \centering
	    \begin{subfigure}
		         \centering
		         \includegraphics[scale=1.5]{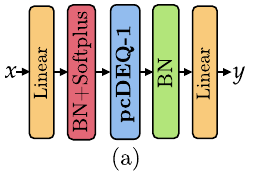}
		\end{subfigure}
	    \begin{subfigure}
		         \centering
		         \includegraphics[scale=1.5]{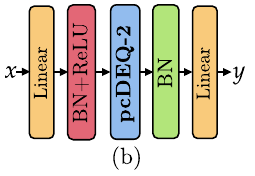}
		     \end{subfigure}
	    \caption{Architectures of pcDEQ models with single linear pcDEQ layer. Subfigure (a) presents architecture with pcDEQ-1 layer and (b) with pcDEQ-2 layer.}
	    \label{fig:fc_archl}
	\end{figure}

Architectures of pcDEQ models with a single linear pcDEQ layer for two variants of pcDEQ layers in the sense of Definition \ref{pcDEQ_def} are presented in Figure \ref{fig:fc_archl}. As in standard DEQ models, the first and last linear layer is an explicit layer.

\subsection{Architecture with Single pcDEQ Convolutional Layer}
\begin{figure}[h!]
	    \centering
	    \begin{subfigure}
		         \centering
		         \includegraphics[scale=1.5]{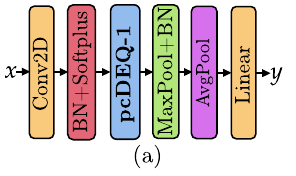}
		     \end{subfigure}
	     \begin{subfigure}
		         \centering
		         \includegraphics[scale=1.5]{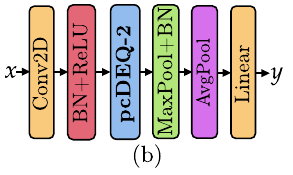}
		     \end{subfigure}
	    \caption{Architecture of pcDEQ models with single convolutional pcDEQ layer. Subfigure (a) presents architecture with pcDEQ-1 layer and (b) with pcDEQ-2 layer.}
	    \label{fig:sc_archl}
	\end{figure}

Architectures of pcDEQ models with a single convolutional pcDEQ layer are similar to those with linear layers in the previous subsection. Figure \ref{fig:sc_archl} shows the architectures used in the experiments. The max. pooling layers have a kernel of size $3 \times 3$ with padding $p=1$ and stride $s=1$. The avg. pooling layer has kernel of size $8 \times 8$ with padding $p=0$ and stride $s=8$.

\subsection{Architecture with Multiple pcDEQ Convolutional Layers}
\begin{figure}[h!]
	    \centering
	    \begin{subfigure}
		         \centering
		         \includegraphics[scale=1.5]{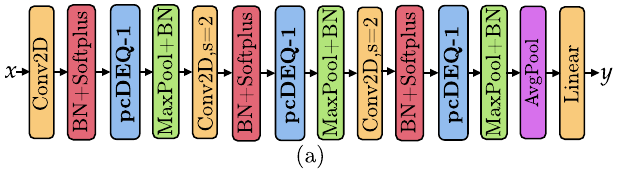}
		     \end{subfigure}
	     \begin{subfigure}
		         \centering
		         \includegraphics[scale=1.5]{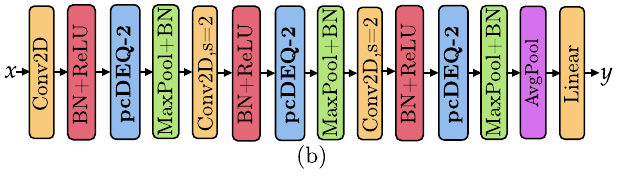}
		     \end{subfigure}
	    \caption{Architecture of pcDEQ models with three convolutional pcDEQ layer. Subfigure (a) presents architecture with pcDEQ-1 layers and (b) with pcDEQ-2 layers.}
	    \label{fig:mc_archl}
	\end{figure}

In practice, larger DEQ models with multiple convolutional layers are implemented as single layer fusion of multiple scales \cite{bai2020multiscale}, between which the upsampling and downsampling are performed. The same idea was applied to monDEQ models in the mult-tier architecture. This is a nice engineering idea, but it is complicated in practice. Moreover, computing the fusion of fixed points (for each scale) simultaneously is difficult to analyze and can involve an increased computational cost. In this work, we take another approach that combines implicit DEQ layers with explicit downsampling layers between them. A similar approach was used in \cite{xie2022optimization}, which investigated the idea of having multiple implicit layers instead of one. It should be noted that the explicit downsampling layers are unconstrained, such as the first and last layers. The proposed architecture is shown in Figure \ref{fig:mc_archl}. The max. pooling layers have a kernel of size $3 \times 3$ with padding $p=1$ and stride $s=1$. The avg. pooling layer has kernel of size $4 \times 4$ with padding $p=0$ and stride $s=4$.

\section{Experimental Details and Hyperparameters}\label{hyper}
In our experiments, we used three commonly known computer vision datasets: MNIST, SVHN, and CIFAR-10. MNIST dataset consists 70,000 grayscale handwritten digit images. SVHN dataset consists of 99,289 RGB digit images from house numbers. CIFAR-10 consists of 60,000 RGB images of 10 classes. For the MNIST dataset, the images have dimensions of $28 \times 28$ pixels, and for the SVHN and CIFAR-10 $32 \times 32$ pixels. The statistics of the datasets are shown in Table \ref{tab:dataset_statistics}.

\begin{table}[!ht]
\centering
\caption{Dataset statistics.}
\label{tab:dataset_statistics}
\begin{tabular}{@{}lcc@{}}
\toprule
\textbf{Dataset} & \textbf{\#Train examples} & \textbf{\#Test examples} \\ \midrule
MNIST & 60,000 & 10,000 \\
SVHN & 73,257 & 26,032 \\
CIFAR-10 & 50,000 & 10,000 \\ \bottomrule
\end{tabular}
\end{table}

As mentioned in the paper, to compute fixed point in pcDEQ layers, the standard fixed point iteration was used. We especially do not use any more sophisticated iterative methods such as Anderson acceleration or Broyden's method, because for such methods the guarantees of convergence to fixed point for PC mappings have not been proved in the literature. For all networks, the AdamW \cite{loshchilov2017decoupled} optimizer was used with a batch size of 64. All other hyperparameters for each dataset and architecture are shown in Tables \ref{tab:mnist_hyper}, \ref{tab:svhn_hyper} and \ref{tab:cifar10_hyper}. The expeeriments with pcDEQs were run five times.

\begin{table}[h!]
	\centering
	\caption{MNIST hyperparameters.}
	\label{tab:mnist_hyper}
	\begin{tabular}{@{}lcccccc@{}}
		\toprule
		\textbf{Method} & \textbf{\begin{tabular}[c]{@{}c@{}}Number of \\ channels\end{tabular}} & \textbf{Epochs} & \textbf{LR} & \textbf{\begin{tabular}[c]{@{}c@{}}LR decay\\ steps\end{tabular}} & \textbf{\begin{tabular}[c]{@{}c@{}}LR decay \\ factor\end{tabular}} & \textbf{WD} \\ \midrule
		pcDEQ-1-L-ReLU6 & 80 & 40 & 0.001 & 30 & 0.1 & 0.02 \\
		pcDEQ-1-L-Tanh & 80 & 40 & 0.001 & 30 & 0.1 & 0.02 \\
		pcDEQ-1-L-Softsign & 80 & 40 & 0.001 & 30 & 0.1 & 0.02 \\
		pcDEQ-2-L-Sigmoid & 80 & 40 & 0.001 & 30 & 0.1 & 0.02 \\
		pcDEQ-1-SC-ReLU6 & 82 & 40 & 0.0007 & 30 & 0.1 & 0.02 \\
		pcDEQ-1-SC-Tanh & 82 & 40 & 0.0007 & 30 & 0.1 & 0.02 \\
		pcDEQ-1-SC-Softsign & 82 & 40 & 0.0007 & 30 & 0.1 & 0.02 \\
		pcDEQ-2-SC-Sigmoid & 82 & 40 &  0.0002 & 30 & 0.1 & 0.02 \\
		pcDEQ-1-MC-ReLU6 & 12,24,48 & 40 & 0.0005 & 30 & 0.1 & 0.015 \\
		pcDEQ-1-MC-Tanh & 12,24,48 & 40 & 0.0005 & 30 & 0.1 & 0.015 \\
		pcDEQ-1-MC-Softsign & 12,24,48 & 40 & 0.0005 & 30 & 0.1 & 0.015 \\
		pcDEQ-2-MC-Sigmoid & 12,24,48 & 40 & 0.0002 & 30 & 0.1 & 0.015 \\ \bottomrule
	\end{tabular}
\end{table}

\begin{table}[H]
	\centering
	\caption{SVHN hyperparameters.}
	\label{tab:svhn_hyper}
	\begin{tabular}{@{}lcccccc@{}}
		\toprule
		\textbf{Method} & \textbf{\begin{tabular}[c]{@{}c@{}}Number of \\ channels\end{tabular}} & \textbf{Epochs} & \textbf{LR} & \textbf{\begin{tabular}[c]{@{}c@{}}LR decay\\ steps\end{tabular}} & \textbf{\begin{tabular}[c]{@{}c@{}}LR decay \\ factor\end{tabular}} & \textbf{WD} \\ \midrule
		pcDEQ-1-SC-ReLU6 & 125 & 80 & 0.0007 & 70 & 0.1 & 0.02 \\
		pcDEQ-1-SC-Tanh & 125 & 80 & 0.0007 & 70 & 0.1 & 0.02 \\
		pcDEQ-1-SC-Softsign & 125 & 80 & 0.0007 & 70 & 0.1 & 0.02 \\
		pcDEQ-2-SC-Sigmoid & 125 & 80 & 0.0005 & 70 & 0.1 & 0.02 \\
		pcDEQ-1-MC-ReLU6 & 20,50,80 & 50 & 0.0005 & 40 & 0.1 & 0.015 \\
		pcDEQ-1-MC-Tanh & 20,50,80 & 50 & 0.0005 & 40 & 0.1 & 0.015 \\
		pcDEQ-1-MC-Softsign & 20,50,80 & 50 & 0.0005 & 40 & 0.1 & 0.015 \\
		pcDEQ-2-MC-Sigmoid & 20,50,80 & 50 & 0.0002 & 40 & 0.1 & 0.015 \\ \bottomrule
	\end{tabular}
\end{table}

\begin{table}[H]
	\centering
	\caption{CIFAR-10 hyperparameters.}
	\label{tab:cifar10_hyper}
	\begin{tabular}{@{}lcccccc@{}}
		\toprule
		\textbf{Method} & \textbf{\begin{tabular}[c]{@{}c@{}}Number of \\ channels\end{tabular}} & \textbf{Epochs} & \textbf{LR} & \textbf{\begin{tabular}[c]{@{}c@{}}LR decay\\ steps\end{tabular}} & \textbf{\begin{tabular}[c]{@{}c@{}}LR decay \\ factor\end{tabular}} & \textbf{WD} \\ \midrule
		pcDEQ-1-SC-ReLU6 & 125 & 80 & 0.0005 & 70 & 0.1 & 0.02 \\
		pcDEQ-1-SC-Tanh & 125 & 80 & 0.0005 & 70 & 0.1 & 0.02 \\
		pcDEQ-1-SC-Softsign & 125 & 80 & 0.0005 & 70 & 0.1 & 0.02 \\
		pcDEQ-2-SC-Sigmoid & 125 & 80 & 0.0002 & 70 & 0.1 & 0.02 \\
		pcDEQ-1-MC-ReLU6 & 20,50,80 & 50 & 0.0007 & 40 & 0.1 & 0.015 \\
		pcDEQ-1-MC-Tanh & 20,50,80 & 50 & 0.0007 & 40 & 0.1 & 0.015 \\
		pcDEQ-1-MC-Softsign & 20,50,80 & 50 & 0.0007 & 40 & 0.1 & 0.015 \\
		pcDEQ-2-MC-Sigmoid & 20,50,80 & 50 & 0.0002 & 40 & 0.1 & 0.015 \\ 
		pcDEQ-1-MC-ReLU6* & 100,120,140 & 120 & 0.0007 & 100 & 0.1 & 0.02 \\
		pcDEQ-1-MC-Tanh* & 100,120,140 & 120 & 0.0007 & 100 & 0.1 & 0.02 \\
		pcDEQ-1-MC-Softsign* & 100,120,140 & 120 & 0.0007 & 100 & 0.1 & 0.02 \\
		pcDEQ-2-MC-Sigmoid* & 100,120,140 & 120 & 0.0002 & 100 & 0.1 & 0.02 \\ \bottomrule
	\end{tabular}
\end{table}

\clearpage
\section{Additional Figures}\label{add_results}
\subsection{MNIST}
\begin{figure*}[h!]
	     \centering
	     \begin{subfigure}
		         \centering
		         \includegraphics[scale=0.375]{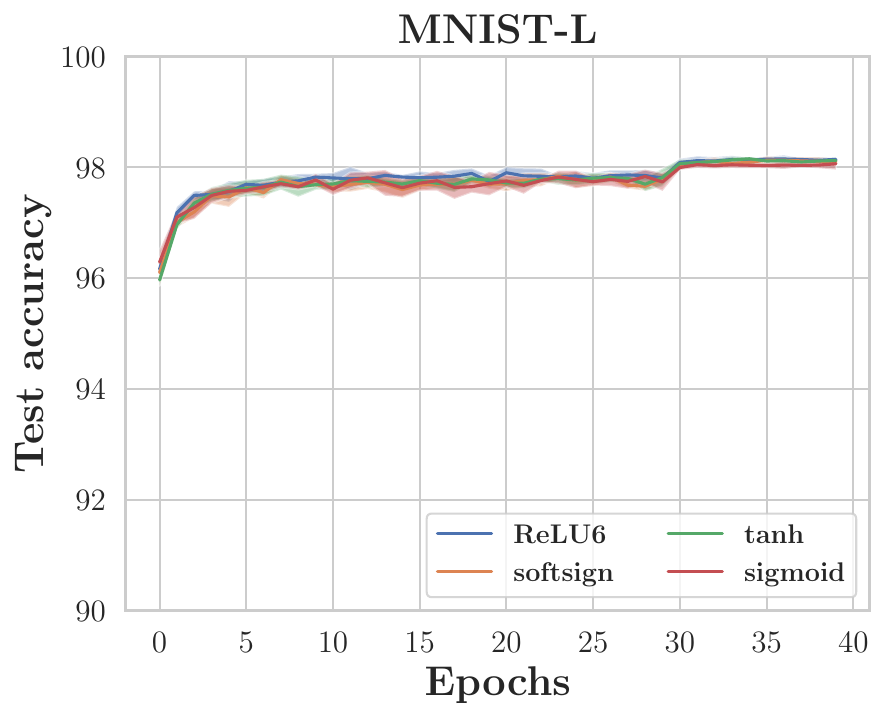}
		     \end{subfigure}
	     \hfill
	     \begin{subfigure}
		         \centering
		         \includegraphics[scale=0.375]{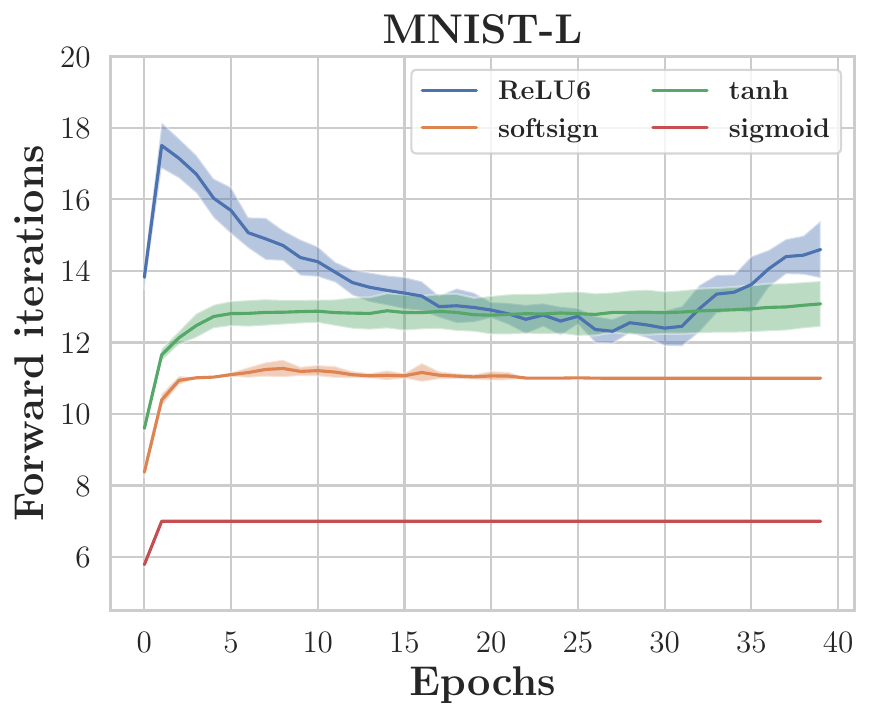}
		     \end{subfigure}
	     \hfill
	     \begin{subfigure}
		         \centering
		         \includegraphics[scale=0.375]{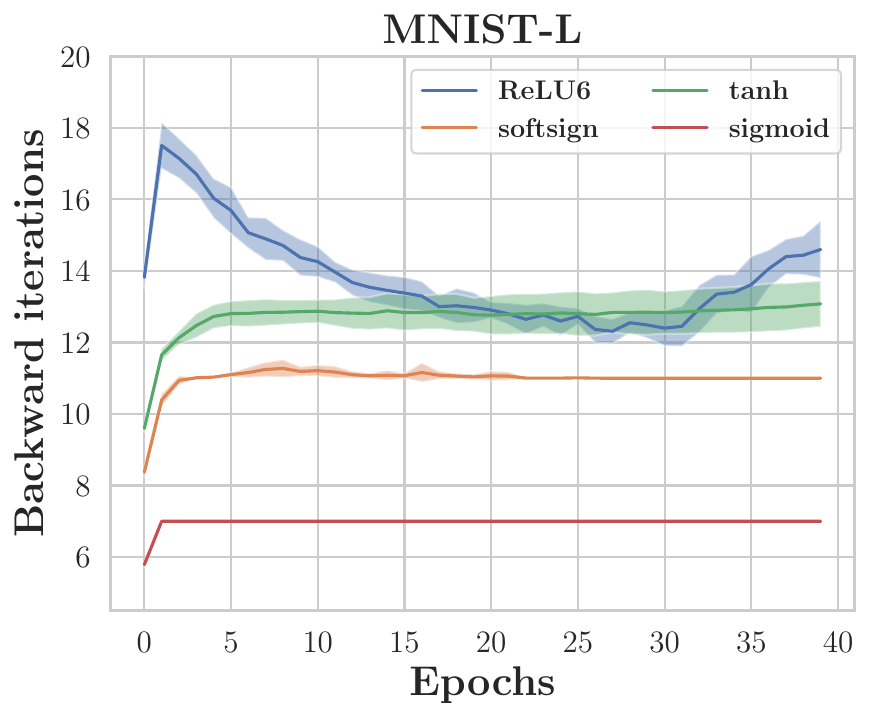}
		     \end{subfigure}
	    \begin{subfigure}
		         \centering
		         \includegraphics[scale=0.37]{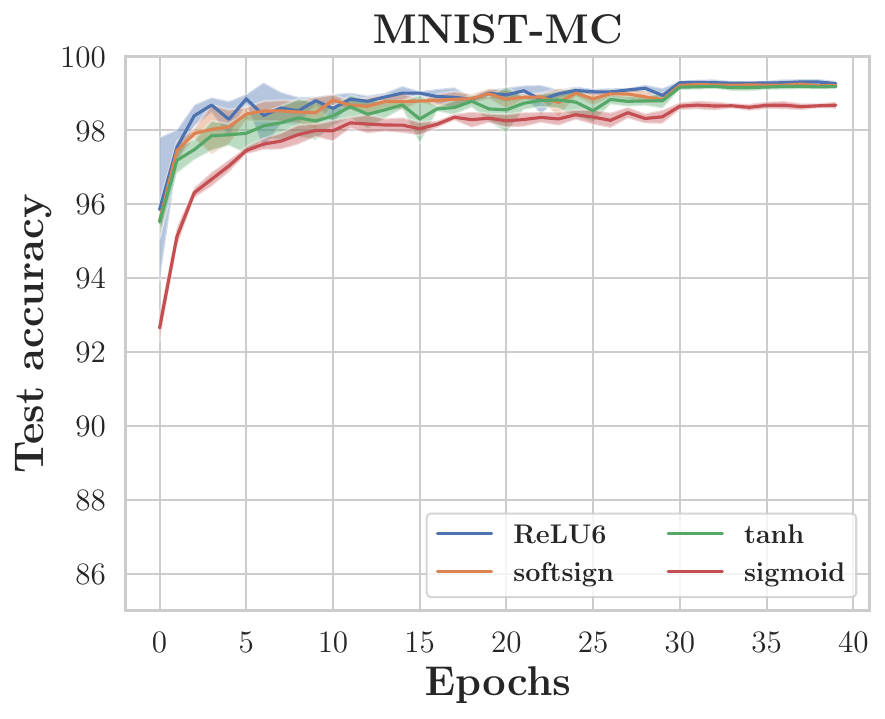}
		     \end{subfigure}
	     \hfill
	     \begin{subfigure}
		         \centering
		         \includegraphics[scale=0.37]{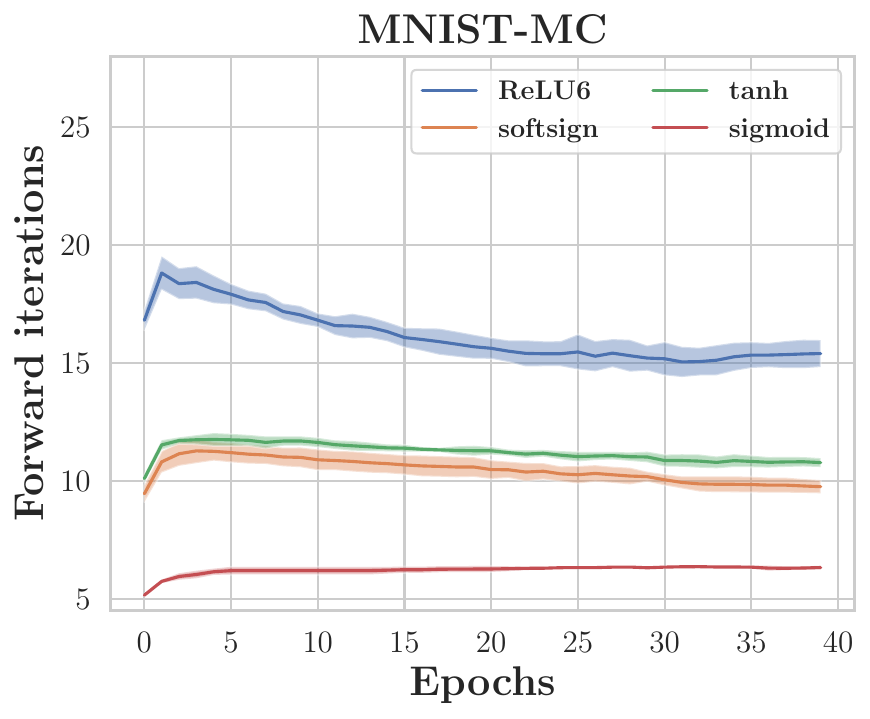}
		     \end{subfigure}
	     \hfill
	     \begin{subfigure}
		         \centering
		         \includegraphics[scale=0.37]{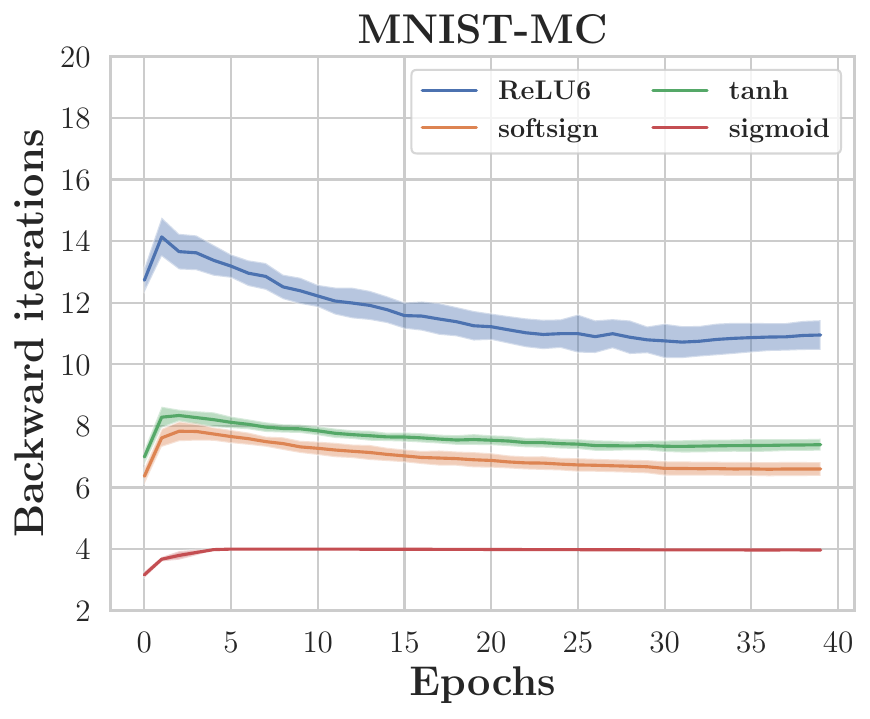}
		     \end{subfigure}
	
	     \label{fig:mnist_additional}
	     \caption{Test accuracies, number of fixed point iterations in forward and backward passes during training for pcDEQ models with single linear pcDEQ layer and three convolutional pcDEQ layers (average of forward and backward passes of three pcDEQ layers) on MNIST dataset over five experiment runs.}
	\end{figure*}

\subsection{SVHN}

\begin{figure*}[h]
	     \centering
	     \begin{subfigure}
		         \centering
		         \includegraphics[scale=0.37]{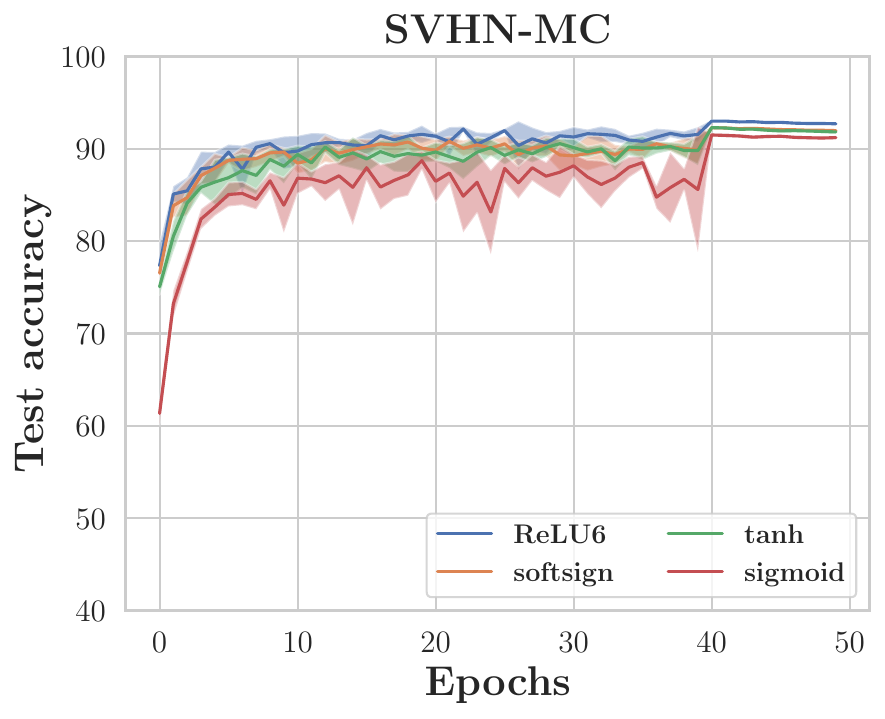}
		     \end{subfigure}
	     \hfill
	     \begin{subfigure}
		         \centering
		         \includegraphics[scale=0.37]{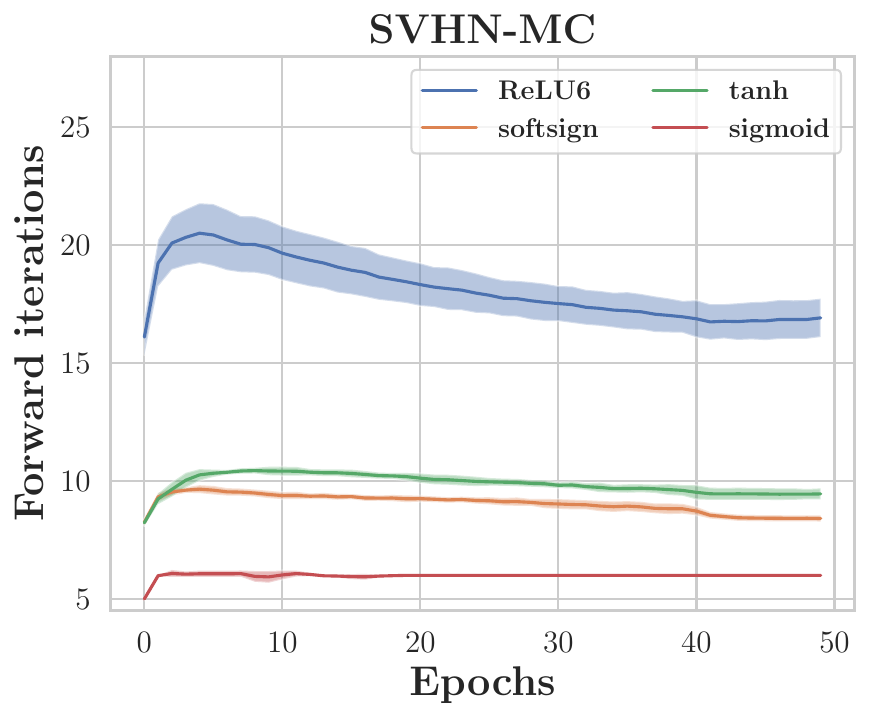}
		     \end{subfigure}
	     \hfill
	     \begin{subfigure}
		         \centering
		         \includegraphics[scale=0.37]{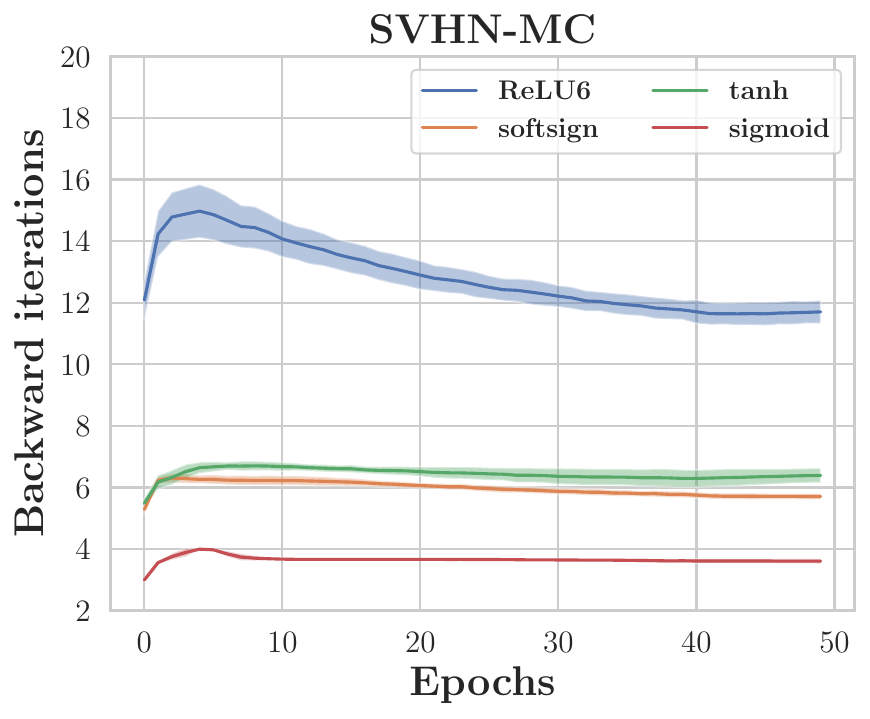}
		     \end{subfigure}
	    \label{}
	    \caption{Test accuracies, number of fixed point iterations in forward and backward passes (average of three pcDEQ layers) during training for pcDEQ models with multiple convolutional pcDEQ  layers on SVHN dataset over five experiment runs.}
	\end{figure*}

\clearpage
\subsection{CIFAR-10}
\begin{figure*}[h!]
	    \begin{subfigure}
		         \centering
		         \includegraphics[scale=0.37]{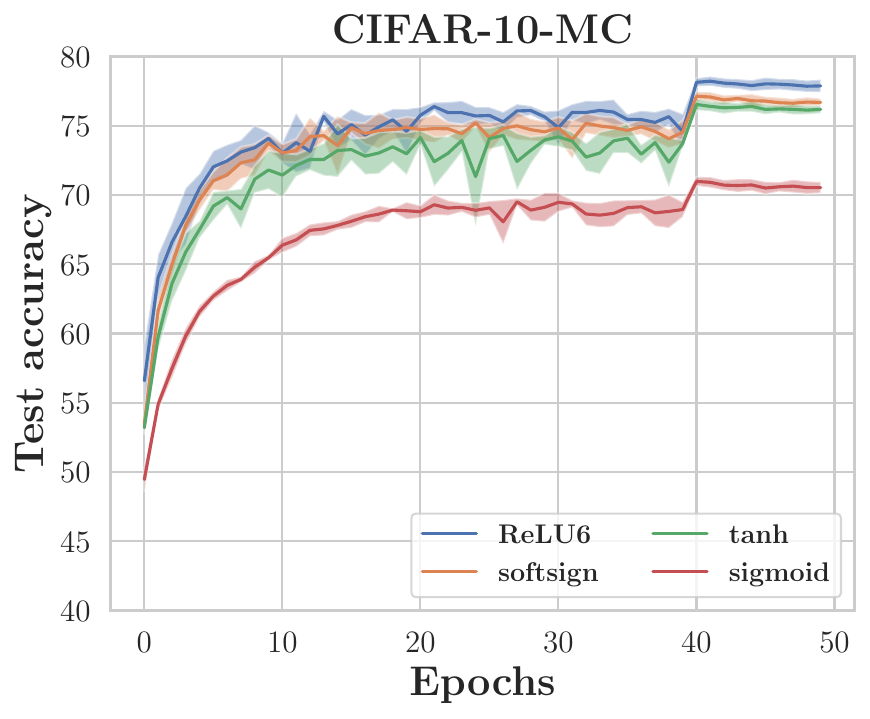}
		     \end{subfigure}
	     \hfill
	     \begin{subfigure}
		         \centering
		         \includegraphics[scale=0.37]{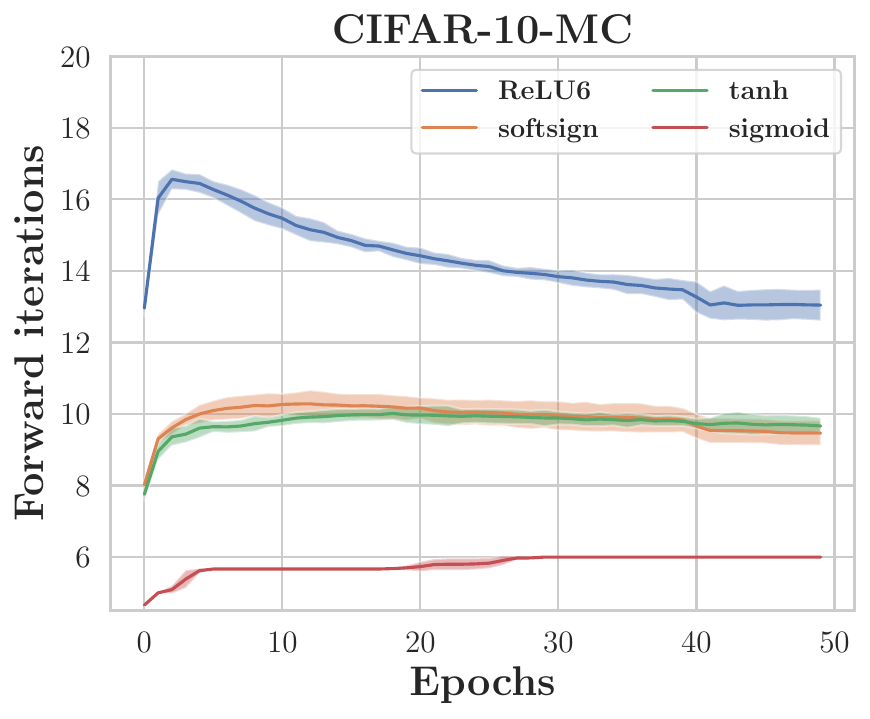}
		     \end{subfigure}
	     \hfill
	     \begin{subfigure}
		         \centering
		         \includegraphics[scale=0.37]{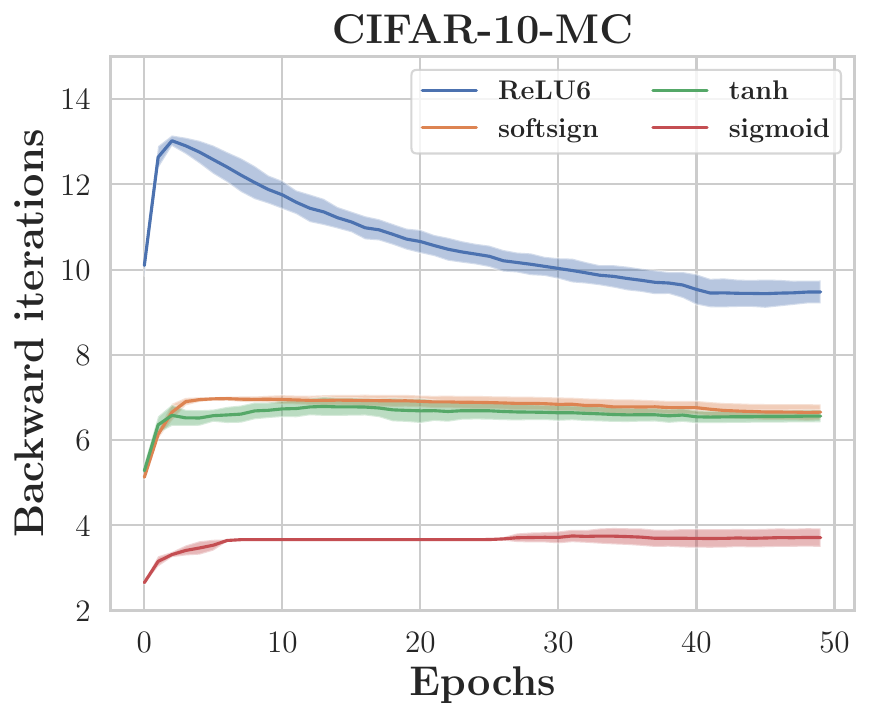}
		     \end{subfigure}
	     \begin{subfigure}
		         \centering
		         \includegraphics[scale=0.37]{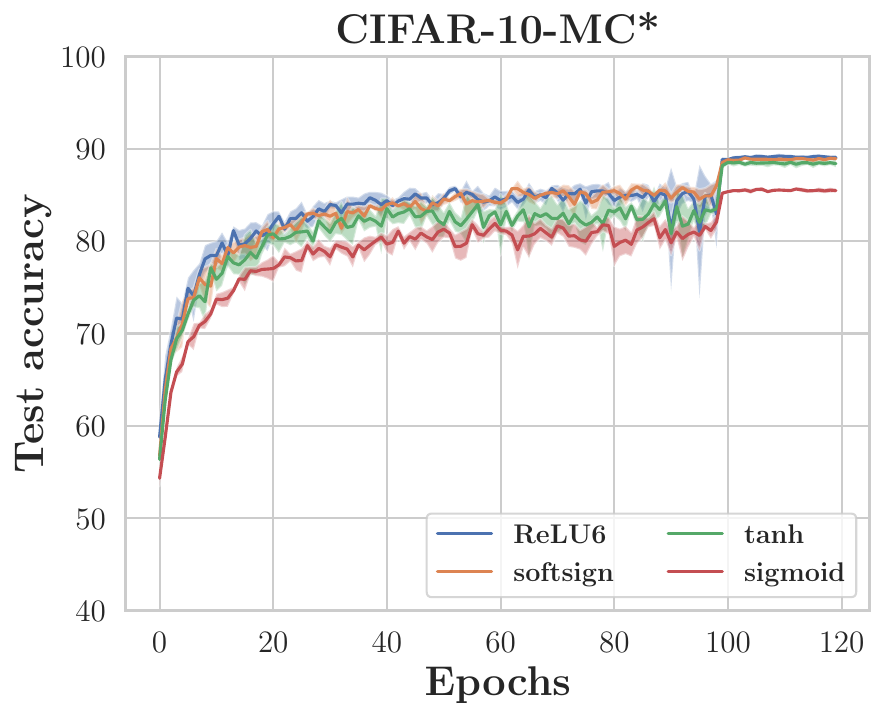}
		     \end{subfigure}
	     \hfill
	     \begin{subfigure}
		         \centering
		         \includegraphics[scale=0.37]{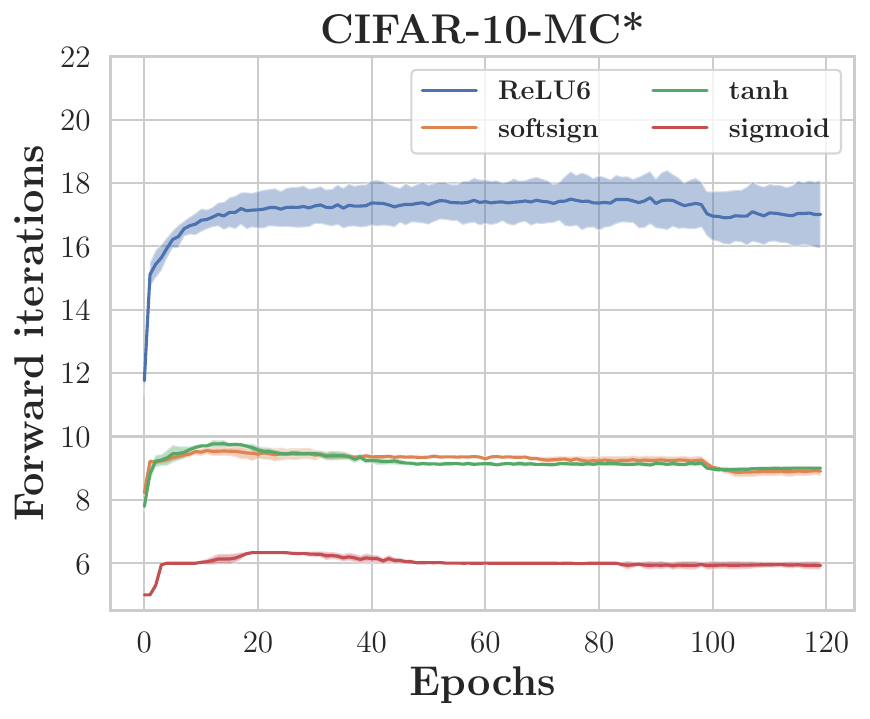}
		     \end{subfigure}
	     \hfill
	     \begin{subfigure}
		         \centering
		         \includegraphics[scale=0.37]{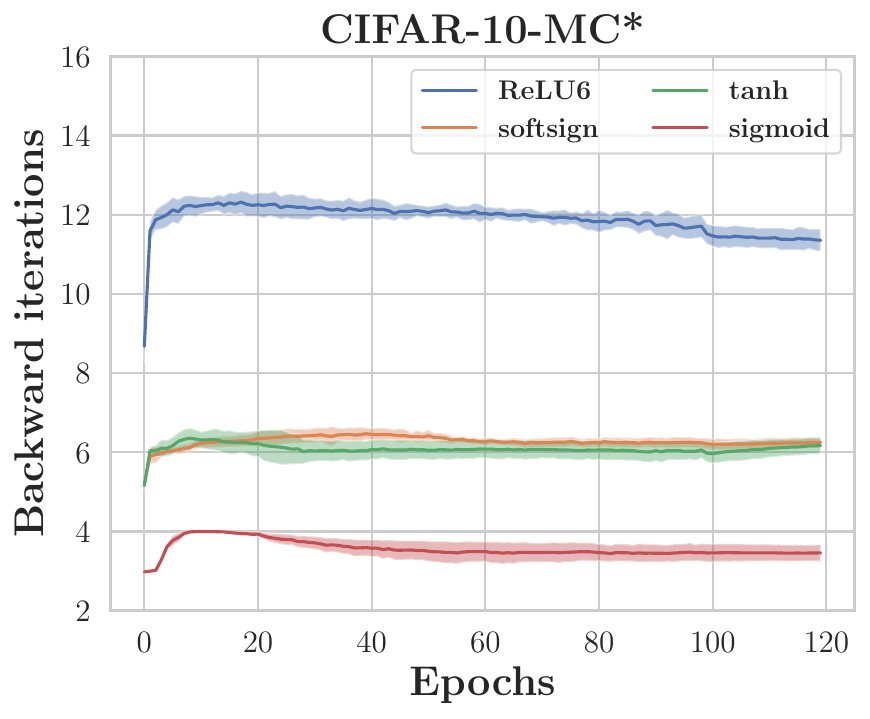}
		     \end{subfigure}
	     \label{}
	     \caption{Test accuracies, number of fixed point iterations in forward and backward passes (average of three pcDEQ layers) during training for pcDEQ models with three convolutional pcDEQ layers with and without data augmentation on CIFAR-10 dataset over five experiment runs.}
	\end{figure*}

\end{document}